\documentclass{article}

\PassOptionsToPackage{numbers, compress}{natbib}

 \usepackage[main, final]{neurips_2026}

\usepackage[utf8]{inputenc} 
\usepackage[T1]{fontenc}    
\usepackage{xurl}
\usepackage{hyperref}       
\usepackage{url}            
\usepackage{booktabs}       
\usepackage{amsfonts}       
\usepackage{nicefrac}       
\usepackage{microtype}      
\usepackage{xcolor}         

\usepackage{amsmath}
\usepackage{wrapfig}
\usepackage{enumitem}
\usepackage{adjustbox}
\usepackage{colortbl}
\usepackage{array,multirow,graphicx}
\usepackage[table]{xcolor}
\usepackage{amssymb}
\usepackage[most]{tcolorbox}
\tcbuselibrary{theorems}
\usepackage{caption}
\usepackage{fontawesome5}

\newcommand{\setB}[0]{\mathcal{B}}
\newcommand{\setD}[0]{\mathcal{D}}
\newcommand{\setL}[0]{\mathcal{L}}

\newcommand{\vv}[0]{\boldsymbol{v}}

\newcommand{\fLLM}[0]{f_{\scriptscriptstyle\mathrm{LLM}}}
\newcommand{\gLLM}[0]{g_{\scriptscriptstyle\mathrm{LLM}}}
\newcommand{\fRET}[0]{f_{\mathrm{ret}}}
\newcommand{\simscore}[0]{\mathrm{cos}}
\newcommand{\ProjRET}[0]{\mathrm{Proj}_{\mathrm{ret}}}
\newcommand{\Projtau}[0]{\mathrm{Proj}_{\tau}}
\newcommand{\Projs}[0]{\mathrm{Proj}_{s}}
\newcommand{\thetaLLM}[0]{\theta_{\scriptscriptstyle\mathrm{LLM}}}
\newcommand{\thetaRET}[0]{\theta_{\mathrm{ret}}}

\newcommand{\tagtoken}[2][]{\texttt{<}\mathtt{#2}_{#1}\texttt{>}}

\definecolor{darkgreen}{RGB}{14,135,67} 
\definecolor{lightgray}{RGB}{245,245,245}

\definecolor{thought_color}{HTML}{336E58}
\definecolor{subquery_color}{HTML}{055A9C}
\definecolor{answer_color}{HTML}{CF1F30}

\definecolor{darkred}{HTML}{D5424E}
\definecolor{darkblue}{HTML}{4E62AA}

\definecolor{best_color}{HTML}{A5D6A7}
\definecolor{second_best_color}{HTML}{FFF3B0}

\definecolor{lightblue}{HTML}{0071BC}

\hypersetup{colorlinks,allcolors=lightblue, urlcolor=lightblue, citecolor=lightblue}

\def\ie{\emph{i.e.}}

\usepackage{pifont}

\newtcolorbox[
  auto counter,
  number within=section,
  list inside=examplelist
]{promptbox}[2][]{
  enhanced,
  title={\textbf{Prompt~\thetcbcounter.} \textbf{#2}},
  fontupper=\footnotesize,
  fonttitle=\small,
  colback=lightgray,
  colframe=black,
  colbacktitle=black,
  coltitle=white,
  #1
}

\newcommand{\cell}[2]{%
  \setlength{\fboxsep}{1pt}%
  \colorbox{#1}{\rule{0pt}{1.2ex}\strut #2}%
}

\title{LatentRAG: Latent Reasoning and Retrieval\\for Efficient Agentic RAG}

\author{%
  Yijia Zheng \qquad Marcel Worring \\
  University of Amsterdam, Amsterdam, the Netherlands \\
  \texttt{\{y.zheng, m.worring\}@uva.nl} \\\\
  \small\href{https://github.com/zhengyijia/LatentRAG}{\textcolor{black}{\faGithub}} \enspace\url{https://github.com/zhengyijia/LatentRAG}
}

\begin{document}

\maketitle

\begin{abstract}
Single-step retrieval-augmented generation (RAG) provides an efficient way to incorporate external information for simple question answering tasks but struggles with complex questions. Agentic RAG extends this paradigm by replacing single-step retrieval with a multi-step process, in which the large language model (LLM) acts as a search agent that generates intermediate thoughts and subqueries to iteratively interact with the retrieval system. This iterative process incurs substantial latency due to the autoregressive generation of lengthy thoughts and subqueries. To address this limitation, we propose \textbf{LatentRAG}, a novel framework that shifts both reasoning and retrieval from discrete language space to continuous latent space. Unlike existing explicit methods that generate natural language thoughts or subqueries token-by-token, LatentRAG produces latent tokens for thoughts and subqueries directly from the hidden states in a single forward pass. We align LLMs with dense retrieval models in the latent space, enabling retrieval over latent subquery tokens and supporting end-to-end joint optimization. To improve transparency and encourage semantically meaningful latent representations, we incorporate a parallel latent decoding mechanism that translates latent tokens back into natural language. Extensive experiments on seven benchmark datasets show that LatentRAG achieves performance comparable to explicit agentic RAG methods while reducing inference latency by approximately \textbf{90\%}, substantially narrowing the latency gap with traditional single-step RAG. 
\end{abstract}

\section{Introduction}\label{sec:introduction}

\begin{figure*}[t]
\begin{center}
\includegraphics[trim={0cm 0cm 0cm 0cm},clip,width=1.0\linewidth]{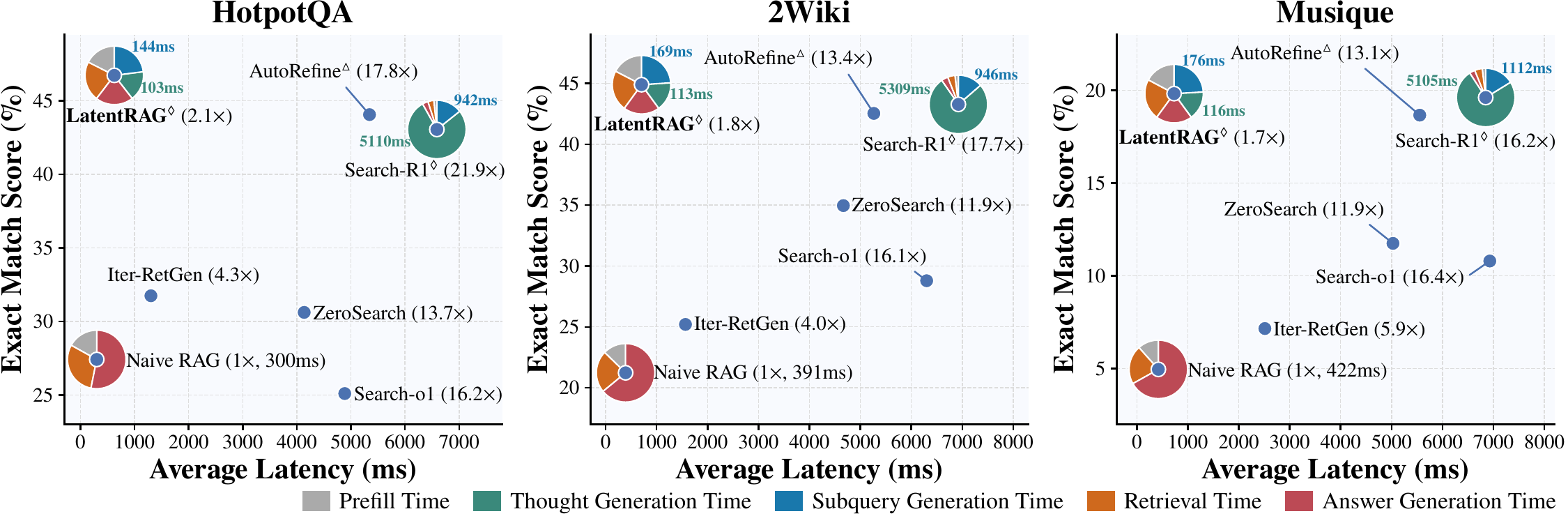}
\end{center}
\vspace{-3pt}
\caption{\textbf{Performance and latency comparison across multi-hop QA datasets.} The parenthesized numbers indicate the latency ratio relative to Naive RAG. LatentRAG achieves performance comparable to competitive agentic RAG methods such as Search-R1 and AutoRefine, while maintaining latency on par with naive single-step RAG. Search-R1 incurs excessive latency from thought and subquery generation, whereas LatentRAG substantially reduces the time spent in these two stages, leading to the efficiency gains. Detailed stage-wise latency breakdowns are provided in Appendix~\ref{app:latency_comparison}.}
\label{fig:performance_stage_inference_time}
\end{figure*}

Large language models (LLMs) have demonstrated strong capabilities in answering complex questions \citep{lai2024large, wang2025asurvey, singhal2025toward}, but these capabilities are fundamentally bounded by their static internal knowledge \citep{wang2025survey, wang2024factuality}. Solely relying on internal knowledge limits their performance on questions that require up-to-date information or proprietary knowledge \citep{wang2024knowledge, wang2025bring} and increases the risk of hallucinations \citep{ji2023survey, huang2025survey}.  To improve both the factuality and transparency of LLM-generated outputs, retrieval-augmented generation (RAG) \citep{lewis2020retrieval, guu2020retrieval} retrieves question-relevant information from an external retrieval system to augment the LLM inputs \citep{gao2023retrieval, peng2025graph}. Traditional RAG methods provide an efficient way to access external knowledge, but their single-step retrieval design limits their effectiveness on complex questions that require iterative reasoning and retrieval \citep{trivedi2023interleaving, singh2025agentic}.

Motivated by the success of tool-using LLM agents \citep{yao2023react, schick2023toolformer}, recent agentic RAG approaches \citep{li2025search, jin2025search} replace traditional single-step retrieval with a multi-step agentic search process that alternates between generation and retrieval. In this process, the LLM acts as a search agent and iteratively decides what to retrieve. At each iteration, the agent generates a thought via chain-of-thought (CoT) reasoning \citep{wei2022chain} and then produces the next action, which can be either a subquery for the next retrieval step or the final answer. Each generated subquery is used to retrieve relevant documents. Unlike static single-step retrieval in traditional RAG, this multi-step agentic search process enables complex questions to be decomposed and effectively solved step by step \citep{li2025towards, jin2025empirical}. Although agentic RAG methods demonstrate strong performance on tasks with complex questions \citep{singh2025agentic, lin2025comprehensive}, they incur substantial latency due to the additional multi-step interactions \citep{guan2026deeprag, tan2026rag}.

To identify the latency bottlenecks of agentic RAG, we measure the average inference time across different stages for both naive single-step RAG and agentic RAG methods. As shown in Fig.~\ref{fig:performance_stage_inference_time}, on multi-hop question answering (QA) datasets, the total inference time of a representative agentic RAG method, Search-R1 \citep{jin2025search}, requires 16–22$\times$ the inference time of naive RAG. This overhead is primarily driven by the thought and subquery generation stages, which together account for approximately 90\% of the total latency. Both stages involve autoregressive token-by-token generation of long outputs, where each output token depends on previously generated tokens, leading to multiple sequential LLM forward passes with limited parallelism. In contrast, prefill, retrieval, and final answer generation take far less time than the other two stages. The inference time comparison indicates that the latency bottlenecks of agentic RAG lie in the thought and subquery generation stages.

To reduce the thought and subquery generation latency in agentic RAG, we draw inspiration from another technique called latent reasoning. Latent reasoning \citep{hao2025training, chen2025reasoning} is an efficient reasoning paradigm that performs reasoning within the continuous hidden states of the LLM, also referred to as latent tokens, without explicitly generating discrete language tokens. Compared to explicit reasoning, latent reasoning avoids allocating computation to non-semantic tokens that are produced solely for linguistic fluency \citep{cheng2024compressed, hao2025training}. Furthermore, continuous latent tokens allow the LLM to directly generate high-level semantic representation, avoiding the inefficiency of explicit token-by-token generation and thereby enabling more parallelizable computation \citep{zhu2025survey, barrault2024large, tack2025llm}. Although latent reasoning offers a promising avenue for enhancing reasoning efficiency, its application to agentic RAG remains unexplored.

In this work, we pioneer the integration of latent reasoning into the agentic RAG paradigm and, more importantly, propose a latent retrieval mechanism. Unlike generation-only tasks studied in prior work on latent reasoning \citep{hao2025training, goyal2024think}, agentic RAG requires the LLM to emit explicit subquery tokens to invoke external retrieval. This explicit token generation not only incurs significant decoding overhead but also prevents gradient propagation, thereby hindering direct optimization of the LLM using retrieval signals. To overcome these limitations, we investigate whether latent tokens generated by an LLM can effectively serve as subqueries for retrieval. This introduces \textbf{two challenges}. \textbf{(1) Data scarcity:} Training retrieval models typically requires large-scale paired data, often comprising hundreds of millions of query–document pairs \citep{zhang2025qwen3, wang2022text}. In contrast, agentic RAG systems are commonly developed under a training setup that provides only tens of thousands of question–answering pairs, without explicit supervision on the ground-truth documents for intermediate subqueries \citep{jin2025search, shi2025search}. This data scarcity makes it difficult to learn effective retrieval capability using conventional training paradigms for retrieval models. \textbf{(2) Transparency:} Latent tokens inherently obscure the intermediate thoughts and subqueries, which is particularly problematic for agentic RAG, as lengthy and redundant retrieved documents make answer verification and evidence attribution \citep{qi2024model, batista2025safe} time-consuming without explicit intermediate steps.

To address the aforementioned challenges, we introduce \textbf{LatentRAG}, an efficient agentic RAG framework that conducts reasoning and retrieval in the latent space. Specifically, we feed a sequence of special thought and subquery tokens into the LLM and use the corresponding last hidden states as latent thought and subquery tokens, respectively. These latent tokens are obtained in a single forward pass, enabling parallel computation and avoiding the inefficiency of autoregressive generation. To address challenge (1), we align the LLM with a pretrained dense retrieval model in the latent space. The latent subquery tokens are used as inputs to the retrieval model to generate latent subquery embeddings. We then minimize the KL divergence between the similarity distribution over documents induced by latent subquery embeddings and that induced by natural language subquery embeddings. This design enables fully differentiable end-to-end joint optimization of the LLM and the retrieval model. To address challenge (2) and encourage the latent tokens to capture meaningful semantics, we incorporate a parallel latent decoding mechanism that converts latent tokens into natural language thoughts and subqueries. During inference, this latent decoding process is optional, enabling a trade-off between transparency and efficiency. Since this latent decoding process depends only on the latent tokens, all thoughts and subqueries across different steps can be decoded in parallel, reducing the latency of the decoding process. Our main contributions are summarized as follows:
\begin{itemize}[leftmargin=*, topsep=1mm]
\item We introduce LatentRAG, a novel agentic RAG framework that performs reasoning and retrieval in the latent space, reducing the latency overhead of explicit thought and subquery generation.
\item We propose a latent-space alignment objective that jointly optimizes the LLM and retrieval model, enabling latent tokens to serve as effective retrieval queries while supporting end-to-end training.
\item We incorporate a parallel decoding mechanism that translates latent tokens into explicit thoughts and subqueries, improving transparency while remaining more efficient than explicit agentic RAG.
\end{itemize}

Extensive experiments on seven benchmark datasets show that LatentRAG achieves performance comparable to explicit agentic RAG methods, with relative performance differences of less than 5\%, while significantly reducing latency overhead by approximately \textbf{90\%} on average, approaching the latency of traditional single-step RAG.

\section{Related Work}

\paragraph{Agentic RAG.} Recent advances in RAG have shifted beyond traditional single-step methods \citep{lewis2020retrieval, guu2020retrieval} toward agentic RAG approaches \citep{lin2025comprehensive, li2025towards, singh2025agentic}, which perform multi-step retrieval by iteratively generating intermediate thoughts and subqueries. Early agentic RAG methods \citep{trivedi2023interleaving, jiang2023active, xinjie2025reagent, li2025search} primarily rely on prompting strategies to enable LLMs to interact with retrieval systems. To improve the retrieval ability of LLMs, Self-RAG \citep{asai2024self} and AutoRAG \citep{kim2024autorag} construct synthetic training data from RAG benchmark datasets for supervised finetuning (SFT). Some methods \citep{guan2026deeprag, cheng2024unified, jeong2024adaptive} further introduce mechanisms to balance internal knowledge and external retrieval, enabling LLMs to retrieve only when internal knowledge is insufficient. To mitigate the reliance on supervised training data and promote more flexible search strategies, a growing line of work \citep{jin2025search, chen2025learning, song2025r1, zheng2025deepresearcher} formulates agentic RAG as a Markov decision process, where LLMs learn an optimal decision policy to interact with the retrieval system via reinforcement learning (RL). Recent RL-based approaches further incorporate fine-grained intermediate reward functions \citep{xie2026tips, wu2026hiprag, zhang2026a, zhao2025r} and explore parallel retrieval strategies \citep{zhao2025parallelsearch, tan2026rag, xu2026wideseek}. As discussed in the introduction, all these existing methods require generating long sequences of thoughts and subqueries in the language space, leading to substantial latency. In contrast to existing approaches, we explore performing reasoning and retrieval in the latent space, avoiding long textual thought and subquery generation and achieving significant efficiency gains.

\paragraph{Latent Reasoning.} Latent reasoning \citep{zhu2025survey, yu2026latent} reduces the latency overhead of explicit chain-of-thought (CoT) reasoning \citep{wei2022chain} by operating in the continuous hidden states of LLMs, but existing work primarily focuses on generation-only tasks \citep{goyal2024think, hao2025training} without external retrieval. Early research explores adding filler tokens to enable LLMs to allocate more computation within the hidden states before generating outputs \citep{goyal2024think, pfau2024let}. Coconut \citep{hao2025training} proposes an autoregressive latent reasoning paradigm, where each latent token, \ie, a generated hidden state, is recursively fed back into the LLM to generate the next latent token. While the training process of Coconut is supervised only by the final answer, subsequent methods \citep{shen2025codi, cheng2024compressed, wang2025synadapt, wei2026sim} further leverage information generated by explicit CoT as intermediate supervision to improve the training process. To enhance semantic consistency and address the distributional mismatch between the latent token space and the model input space, recent approaches \citep{zhang2025soft, zhou2026geometry, deng2025latent} constrain latent representations as mixtures of the language token embeddings. Some methods \citep{he2025semcot, xu2025softcot} introduce lightweight assistant models to generate latent tokens, thereby improving efficiency while avoiding disruption to the capabilities of the base LLM. Latent reasoning has been extended to practical applications, including retrieval. CLaRa \citep{he2025clara} leverages latent reasoning to compress retrieved information in single-step RAG, while a concurrent work, LaSER \citep{jin2026laser}, develops a dense retrieval model based on latent reasoning. Despite the rapid advancement of latent reasoning, its application to agentic RAG introduces several challenges as discussed in the introduction, leaving this area largely unexplored. In this paper, we pioneer the integration of latent reasoning into agentic RAG and further propose a latent retrieval mechanism, significantly reducing latency overhead.

\section{Preliminaries}\label{sec:preliminaries}

Following the standard setting in prior RAG research \citep{jin2025search, tan2026rag}, we study the question-answering (QA) task defined as follows. Given a question $q$, the objective is to generate an answer $a$ by retrieving the necessary information from a large corpus $\setD = \{ d_1, d_2, \ldots, d_N \}$, where each $d_i$ represents a document. To simplify notation, for each symbol that represents natural language text (e.g., $d_i$), we use the same symbol to denote its token sequence.

LLMs are widely used for solving the QA task. An LLM maps an input token sequence to an output sequence through two stages: \textit{prefill} and \textit{decoding}. In the prefill stage, all input tokens are processed in parallel to compute the key-value (KV) cache. In the decoding stage, output tokens are generated autoregressively, where each token is produced based on the KV cache of the input tokens and previously generated tokens. Due to autoregressive dependencies, the decoding stage can only generate output tokens in a token-by-token manner, leading to substantial latency for long outputs.

RAG systems augment LLMs with information retrieved from an external retrieval system. Two types of retrieval models are widely adopted \citep{singh2025agentic, fan2024survey}: sparse retrieval models, which rely on exact token-level matches, and dense retrieval models, which encode the query and documents into continuous embeddings and select top-$k$ documents based on cosine similarity. Dense retrieval models capture deeper semantic similarity than sparse retrieval models, leading to superior performance on RAG benchmarks \citep{jin2025flashrag, lyu2025crud}. Thus, in this work, we focus on dense retrieval models.

Agentic RAG methods perform multi-step generation and retrieval, as shown in Fig.~\ref{fig:framework}(a). At each iteration, the LLM generates a reasoning thought and a corresponding subquery, which is then used to retrieve relevant information from an external retrieval system. Formally, at iteration $t$, the historical \textit{interaction trajectory} is denoted as a sequence: 
\begin{gather}\label{eq:agentic_rag_trajectory}
\mathcal{I}_t = ( \tau_0, s_0, c_0, \ldots, \tau_{t-1}, s_{t-1}, c_{t-1} ),
\end{gather}
where $\tau_i$ represents the $i$-th thought, $s_i$ is the $i$-th generated subquery, and $c_i$ comprises the contents of the top-$k$ documents retrieved using $s_i$. Conditioned on the question $q$ and the interaction trajectory $\mathcal{I}_t$, the agent first performs reasoning by producing the next thought $\tau_t$ and subsequently generates the next action output $o_{t}$, denoted jointly as $(\tau_t, o_t) = \gLLM (q, \mathcal{I}_t; \thetaLLM)$, where $\thetaLLM$ represents the parameters of the LLM. The action output $o_t$ is either a new subquery $s_t$ for further retrieval or a final answer $a$, distinguished according to the output text format specified by the particular method.

\section{Methodology}\label{sec:methodology}

\begin{figure*}[t]
\begin{center}
\includegraphics[trim={0cm 0cm 0cm 0cm},clip,width=1.0\linewidth]{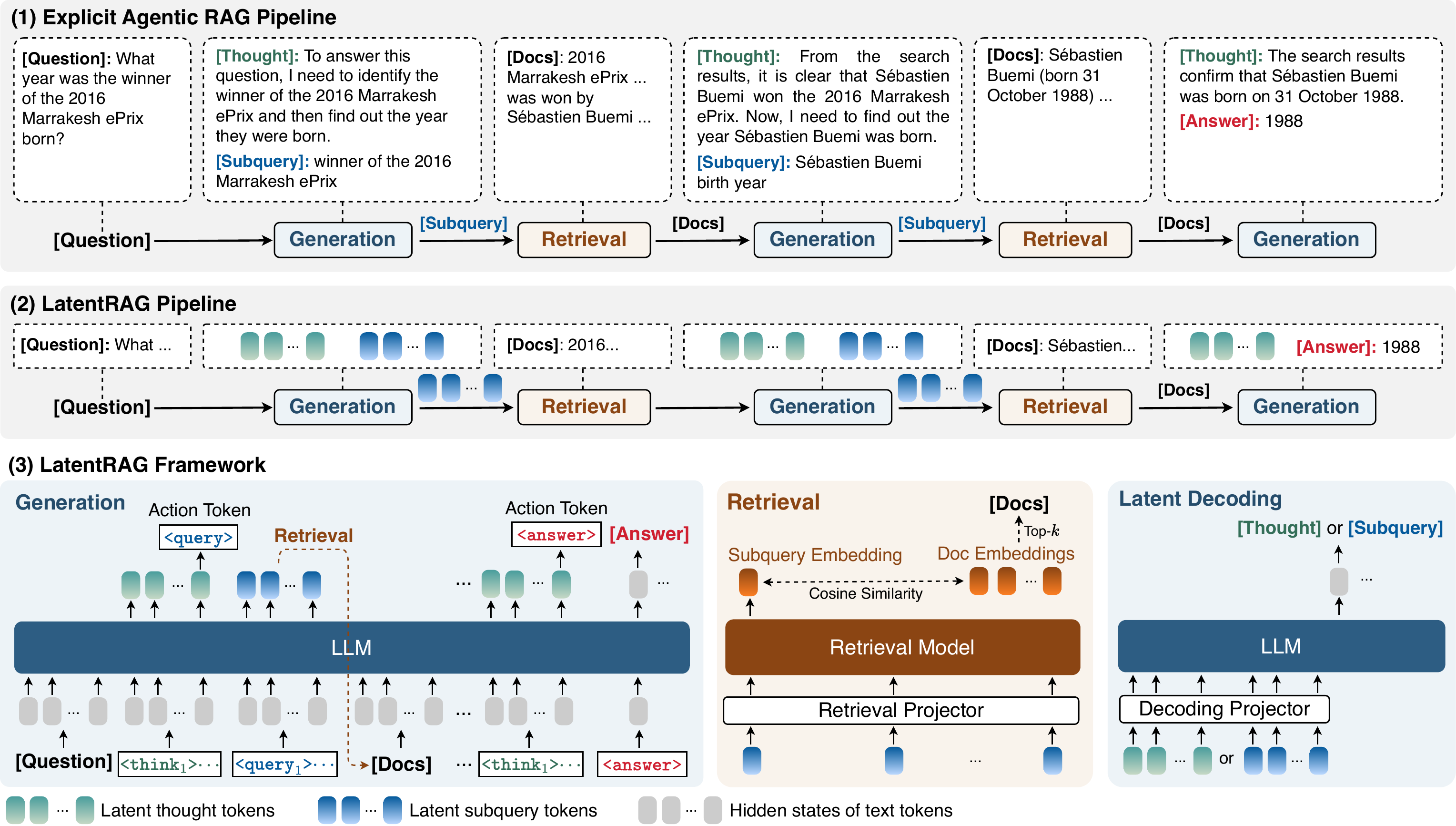}
\end{center}
\caption{(1) Traditional explicit agentic RAG methods alternate between generation and retrieval, producing natural language thoughts and subqueries at each generation step to iteratively retrieve relevant documents. (2) LatentRAG only produces latent thought and subquery tokens at each generation step, and the latent subquery tokens are used for retrieval. (3) LatentRAG contains three components: \textbf{Generation} (Sec.~\ref{sec:generation_with_latent_reasoning}), \textbf{Retrieval} (Sec.~\ref{sec:latent_retrieval}), and \textbf{Latent Decoding} (Sec.~\ref{sec:latent_decoding}).
}
\label{fig:framework}
\end{figure*}

LatentRAG adopts a similar procedure to traditional explicit agentic RAG described in Sec.~\ref{sec:preliminaries}, where the LLM agent iteratively generates thoughts and subqueries, and the subqueries are then used to retrieve relevant information. Unlike explicit agentic RAG methods that generate thoughts and subqueries in the language space, LatentRAG operates in the latent space and only produces latent tokens, \ie, the last-layer hidden states, for thoughts and subqueries (Sec.~\ref{sec:generation_with_latent_reasoning}). The latent subquery tokens are then used as inputs to retrieve relevant documents (Sec.~\ref{sec:latent_retrieval}). To improve transparency, the latent thought and subquery tokens can be decoded into natural language via the latent decoding process (Sec.~\ref{sec:latent_decoding}). The model is trained with a joint objective that combines losses from different components (Sec.~\ref{sec:training_procedure}). The overall framework is shown in Fig.~\ref{fig:framework}.

\subsection{Generation with Latent Tokens}\label{sec:generation_with_latent_reasoning}

We replace the explicit thoughts $\tau_t$ and subqueries $s_t$ in Eq.~\ref{eq:agentic_rag_trajectory} with sequences of special tokens $\tau^{\ell}_t$ and $s^{\ell}_t$, respectively. Here $\tau^{\ell}_t = ( \tagtoken[1]{think}, \ldots,  \tagtoken[m]{think} )$ denotes a sequence of $m$ special thought tokens, and $s^{\ell}_t = ( \tagtoken[1]{query}, \ldots, \tagtoken[n]{query} )$ denotes a sequence of $n$ special subquery tokens. At iteration $t$, the interaction trajectory is denoted as a sequence:
\begin{gather}
\mathcal{I}^{\ell}_t = ( \tau^{\ell}_0, s^{\ell}_0, c_0, \ldots, \tau^{\ell}_{t-1}, s^{\ell}_{t-1}, c_{t-1} ).
\end{gather}
The special tokens serve as latent computation slots, allowing the LLM to allocate additional internal computation without generating explicit natural language thoughts and subqueries. During the prefill stage, the special tokens are processed in parallel, producing their last hidden states $H^{\tau}_{t}$ and $H^{s}_{t}$, which are referred to as latent thought and subquery tokens, respectively.

Given the question $q$ and the interaction trajectory $\mathcal{I}^{\ell}_t$, we append the input with the special thought tokens $\tau^{\ell}_t$ and let the LLM decode an action token from the last latent thought token:
\begin{gather}
\alpha_t = \gLLM( q, \mathcal{I}^{\ell}_t, \tau^{\ell}_{t}; \thetaLLM ),
\end{gather}
where $\alpha_t \in \{ \tagtoken{query}, \tagtoken{answer} \}$ represents whether to proceed with retrieval by generating a subquery or to terminate by producing the final answer. If $\alpha_t = \tagtoken{query}$, we append the special tokens $s^{\ell}_t$ to the input sequence $( q, \mathcal{I}^{\ell}_t, \tau^{\ell}_{t} )$ and let the LLM generate the latent subquery tokens:
\begin{gather}
H^{s}_{t} = \fLLM( s^{\ell}_t; q, \mathcal{I}^{\ell}_t, \tau^{\ell}_{t}, \thetaLLM ),
\end{gather}
where $\fLLM$ denotes a single forward pass of the LLM. The obtained latent subquery tokens $H^{s}_{t}$ are used to retrieve relevant top-$k$ documents, which constitute the retrieved content $c_t$ (described in Sec.~\ref{sec:latent_retrieval}). If $\alpha_t = \tagtoken{answer}$, the model continues to generate the final answer $a$. 

LatentRAG is trained via supervised finetuning using interaction trajectories constructed from existing explicit agentic RAG methods. Specifically, we replace each natural language thought $\tau_{t}$ and subquery $s_{t}$ with the corresponding special token sequences $\tau^{\ell}_{t}$ and $s^{\ell}_t$. The token sequences within these trajectories are formatted according to a predefined prompt template (see Appendix~\ref{app:prompts}) and then provided as input to the LLM. The LLM is optimized to generate the correct action token $\alpha_t$ and the final answer $a$ using the standard cross-entropy loss, denoted as $\setL_{\mathrm{gen}}$.

\subsection{Latent Retrieval}\label{sec:latent_retrieval}

We use the generated latent subquery tokens $H^{s}_{t}$ to retrieve relevant content $c_t$. Since these latent tokens reside in the output space of the LLM and are not directly compatible with the input space of the retrieval model, we add a lightweight projector module $\ProjRET$ to bridge the two spaces. The projector is composed of a bidirectional self-attention layer and a position-wise feed-forward network (FFN) layer. The projected latent subquery tokens are fed into a trainable retrieval model to obtain the latent subquery embedding:
\begin{gather}
\vv_{s^{\ell}_{t}} = \fRET ( \ProjRET (H^{s}_{t}); \thetaRET ).
\end{gather}
Here $\thetaRET$ denotes the parameters of the retrieval model, which are initialized from a pretrained model and will be optimized during the finetuning process. Since ground truth documents are not available in our setting, we train the model to produce latent subquery embeddings that approximate the retrieval behavior induced by the corresponding natural language subqueries. Specifically, each natural language subquery $s_t$ in the trajectory is encoded using a reference retrieval model to produce a reference embedding $\vv_{s_t}'$:
\begin{gather}
\vv_{s_t}' = \fRET(s_t; \thetaRET'),
\end{gather}
where the reference retrieval model is initialized from the same pretrained model as the trainable one, but its parameters $\thetaRET'$ remain frozen during the finetuning process. The reference embeddings are used to retrieve top-$k$ documents from the corpus, which are treated as pseudo-relevant documents.

To learn subquery embeddings that align with relevant documents, a common practice is to use the InfoNCE loss \citep{oord2018representation}, which pulls query embeddings closer to positive documents while pushing them away from negatives. However, in our setting, pseudo-relevant documents are not ground-truth annotations and may contain substantial noise. In addition, unlike large-scale dense retrieval pretraining settings that rely on hundreds of millions of labeled query–document pairs \citep{zhang2025qwen3, wang2022text}, agentic RAG is typically trained with only tens of thousands of samples \citep{jin2025search, shi2025search}. Such data noise and scarcity make the standard InfoNCE objective less well-suited to the agentic RAG setting.

To better leverage the prior knowledge encoded in the pretrained retrieval models, we introduce a retrieval objective based on Kullback–Leibler (KL) divergence. Specifically, for each subquery $s_t$ and each candidate document $d_i$, we compute the following cosine similarity-based probabilities using the reference subquery embedding and the corresponding latent subquery embedding:
\begin{gather}
p_i (s_t) = \frac{ \exp ( \simscore ( \vv_{s_t}', \vv_{d_i} ) / \beta ) }{ \sum_{j=1}^{N_d} \exp ( \simscore ( \vv_{s_t}', \vv_{d_j} ) / \beta ) }, \quad\quad q_i (s_t) = \frac{ \exp ( \simscore ( \vv_{s^{\ell}_{t}}, \vv_{d_i} ) / \beta ) }{ \sum_{j=1}^{N_d} \exp ( \simscore ( \vv_{s^{\ell}_{t}}, \vv_{d_j} ) / \beta ) }, 
\end{gather}
where $\beta$ is the temperature parameter that controls the sharpness of the distribution. $N_d$ is the number of candidate documents, where the candidate set consists of all in-batch pseudo-relevant documents. The retrieval loss function is defined as the KL divergence between both distributions:
\begin{gather}\label{eq:retrieval_loss}
\setL_{\mathrm{ret}} = \frac{1}{\lVert \setB_s \rVert} \sum_{s_t \in \setB_s} \sum_{i=1}^{N_d} p_i (s_t) \log \frac{ p_i (s_t) }{ q_i (s_t) },
\end{gather}
where $\setB_s$ denotes all the subqueries in a training batch. An alternative objective is to directly align $\vv_{s^{\ell}_{t}}$ and $\vv_{s_t}'$ by minimizing cosine distance. However, our ablation experiments show that it yields lower performance compared to the KL objective.

\begin{table}[t]
\small{\caption{\textbf{Overall results with different retrieval models.} \textbf{\textcolor{darkgreen}{Green}} and \textbf{\textcolor{red}{Red}} values respectively indicate relative percentage improvement and degradation compared to the corresponding baseline. \cell{best_color}{Green} and \cell{second_best_color}{yellow} shaded cells indicate the best and second best across all settings. Results show that our method achieves substantial efficiency gains with limited or even no performance degradation.}
\label{tab:main_results}} \centering
\setlength{\tabcolsep}{1pt}
\begin{adjustbox}{max width = 1.0\linewidth}
\begin{tabular}{l|cccccccccccccc|cc}
\toprule
\multirow{2}{*}{\textbf{Methods}} & \multicolumn{2}{c}{\textbf{NQ}} & \multicolumn{2}{c}{\textbf{TriviaQA}} & \multicolumn{2}{c}{\textbf{PopQA}} & \multicolumn{2}{c}{\textbf{HotpotQA}} & \multicolumn{2}{c}{\textbf{2wiki}} & \multicolumn{2}{c}{\textbf{Musique}} & \multicolumn{2}{c|}{\textbf{Bamboogle}} & \multicolumn{2}{c}{\textbf{Average}} \\
& EM(\%)$^{\uparrow}$ & Lat.(ms)$^{\downarrow}$ & EM(\%)$^{\uparrow}$ & Lat.(ms)$^{\downarrow}$ & EM(\%)$^{\uparrow}$ & Lat.(ms)$^{\downarrow}$ & EM(\%)$^{\uparrow}$ & Lat.(ms)$^{\downarrow}$ & EM(\%)$^{\uparrow}$ & Lat.(ms)$^{\downarrow}$ & EM(\%)$^{\uparrow}$ & Lat.(ms)$^{\downarrow}$ & EM(\%)$^{\uparrow}$ & Lat.(ms)$^{\downarrow}$ & EM(\%)$^{\uparrow}$ & Lat.(ms)$^{\downarrow}$ \\ 
\midrule
Direct Infer & 14.76 & 257 & 42.48 & 158 & 14.50 & 175 & 16.68 & 179 & 21.89 & 191 & 3.19 & 239 & 11.20 & 200 & 17.81 & 200 \\
\midrule
\rowcolor{gray!30} \multicolumn{17}{c}{\textbf{Qwen3-Embedding-0.6B}} \\
\midrule
Naive RAG & 25.84 & 452 & 53.32 & 334 & 35.72 & 297 & 27.41 & 300 & 21.23 & 391 & 4.96 & 422 & 11.20 & 314 & 25.67 & 359 \\
Iter-RetGen & 31.33 & 1,884 & 56.91 & 1,241 & 40.29 & 1,297 & 31.74 & 1,303 & 25.20 & 1,567 & 7.16 & 2,513 & 19.20 & 1,461 & 30.26 & 1,609 \\
Search-o1 & 22.33 & 4,178 & 44.53 & 4,102 & 31.25 & 3,422 & 25.09 & 4,886 & 28.79 & 6,301 & 10.80 & 6,928 & 29.60 & 4,741 & 27.48 & 4,937 \\
R1-Searcher & 36.34 & 7,569 & 56.10 & 7,894 & 38.37 & 7,426 & 43.34 & 9,306 & 46.99 & 8,861 & 17.83 & 10,063 & 34.40 & 7,491 & 39.05 & 8,373 \\
ZeroSearch & 34.07 & 4,194 & 54.67 & 3,866 & 40.72 & 3,484 & 30.61 & 4,136 & 34.96 & 4,667 & 11.75 & 5,027 & 32.80 & 4,166 & 34.23 & 4,220 \\
DeepRAG & 33.16 & 2,847 & 55.48 & 2,860 & 39.15 & 2,873 & 33.44 & 3,232 & 43.92 & 3,758 & 13.20 & 4,149 & 32.00 & 2,848 & 35.76 & 3,224 \\
\midrule
Search-R1$^{\lozenge}$ & 43.93 & 3,553 & \textbf{61.70} & 5,198 & \textbf{47.31} & 3,588 & 43.05 & 6,589 & 43.27 & 6,925 & 19.61 & 6,846 & 38.40 & 4,904 & 42.47 & 5,372 \\
\textbf{LatentRAG}$^{\lozenge}$ & \textbf{46.29} & 491\textbf{\textcolor{darkgreen}{(-86.2\%)}} & 59.85 & 478\textbf{\textcolor{darkgreen}{(-90.8\%)}} & 46.63 & 501\textbf{\textcolor{darkgreen}{(-86.0\%)}} & \textbf{46.73} & 626\textbf{\textcolor{darkgreen}{(-90.5\%)}} & \textbf{44.89} & 704\textbf{\textcolor{darkgreen}{(-89.8\%)}} & \textbf{19.82} & 730\textbf{\textcolor{darkgreen}{(-89.3\%)}} & \textbf{40.00} & 623\textbf{\textcolor{darkgreen}{(-87.3\%)}} & \textbf{43.46}\textbf{\textcolor{darkgreen}{(+2.3\%)}} & \textbf{593}\textbf{\textcolor{darkgreen}{(-89.0\%)}} \\
\midrule
AutoRefine$^{\vartriangle}$ & 44.35 & 4,782 & \textbf{63.09} & 4,223 & 47.19 & 4,397 & 44.06 & 5,344 & 42.53 & 5,264 & 18.66 & 5,553 & \textbf{39.20} & 4,224 & 42.73 & 4,827 \\
\textbf{LatentRAG}$^{\vartriangle}$ & \textbf{45.73} & 409\textbf{\textcolor{darkgreen}{(-91.4\%)}} & 60.18 & 400\textbf{\textcolor{darkgreen}{(-90.5\%)}} & \textbf{47.42} & 422\textbf{\textcolor{darkgreen}{(-90.4\%)}} & \textbf{47.16} & 528\textbf{\textcolor{darkgreen}{(-90.1\%)}} & \cellcolor{best_color} \textbf{46.14} & 607\textbf{\textcolor{darkgreen}{(-88.5\%)}} & \textbf{20.73} & 639\textbf{\textcolor{darkgreen}{(-88.5\%)}} & \textbf{39.20} & 581\textbf{\textcolor{darkgreen}{(-86.2\%)}} & \textbf{43.79}\textbf{\textcolor{darkgreen}{(+2.5\%)}} & \textbf{512}\textbf{\textcolor{darkgreen}{(-89.4\%)}} \\
\midrule
\rowcolor{gray!30} \multicolumn{17}{c}{\textbf{e5-base-v2}} \\
\midrule
Search-R1$^{\lozenge}$ & 48.03 & 3,301 & \textbf{64.59} & 5,409 & \textbf{46.02} & 3,403 & 44.71 & 6,263 & \textbf{42.13} & 6,662 & \textbf{20.19} & 6,416 & \textbf{43.20} & 4,709 & \textbf{44.12} & 5,166 \\
\textbf{LatentRAG}$^{\lozenge}$ & \cellcolor{second_best_color} \textbf{49.03} & 453\textbf{\textcolor{darkgreen}{(-86.3\%)}} & 61.30 & 410\textbf{\textcolor{darkgreen}{(-92.4\%)}} & 43.04 & 470\textbf{\textcolor{darkgreen}{(-86.2\%)}} & \textbf{46.41} & 588\textbf{\textcolor{darkgreen}{(-90.6\%)}} & 38.61 & 687\textbf{\textcolor{darkgreen}{(-89.7\%)}} & 18.87 & 656\textbf{\textcolor{darkgreen}{(-89.8\%)}} & 39.20 & 558\textbf{\textcolor{darkgreen}{(-88.2\%)}} & 42.35\textbf{\textcolor{red}{(-4.0\%)}} & \textbf{546}\textbf{\textcolor{darkgreen}{(-89.4\%)}} \\
\midrule
AutoRefine$^{\vartriangle}$ & 48.20 & 4,303 & \cellcolor{best_color} \textbf{65.30} & 3,853 & \textbf{47.12} & 4,135 & 45.13 & 5,133 & 41.90 & 4,695 & \textbf{21.80} & 5,583 & \cellcolor{best_color} \textbf{48.00} & 4,045 & \cellcolor{best_color} \textbf{45.35} & 4,535 \\
\textbf{LatentRAG}$^{\vartriangle}$ & \cellcolor{best_color} \textbf{49.86} & 369\textbf{\textcolor{darkgreen}{(-91.4\%)}} & 62.32 & 370\textbf{\textcolor{darkgreen}{(-90.4\%)}} & 43.88 & 407\textbf{\textcolor{darkgreen}{(-90.2\%)}} & \textbf{46.27} & 464\textbf{\textcolor{darkgreen}{(-91.0\%)}} & \textbf{41.98} & 552\textbf{\textcolor{darkgreen}{(-88.2\%)}} & 20.94 & 548\textbf{\textcolor{darkgreen}{(-90.2\%)}} & 40.80 & 519\textbf{\textcolor{darkgreen}{(-87.2\%)}} & 43.72\textbf{\textcolor{red}{(-3.6\%)}} & \textbf{462}\textbf{\textcolor{darkgreen}{(-89.8\%)}} \\
\midrule
\rowcolor{gray!30} \multicolumn{17}{c}{\textbf{jina-embeddings-v5-text-nano}} \\
\midrule
Search-R1$^{\lozenge}$ & 45.79 & 3,381 & \textbf{63.16} & 5,120 & 46.54 & 3,639 & 44.31 & 6,161 & 42.95 & 6,758 & 20.40 & 6,177 & \cellcolor{second_best_color} \textbf{44.80} & 4,579 & 43.99 & 5,116 \\
\textbf{LatentRAG}$^{\lozenge}$ & \textbf{47.37} & 456\textbf{\textcolor{darkgreen}{(-86.5\%)}} & 61.25 & 394\textbf{\textcolor{darkgreen}{(-92.3\%)}} & \textbf{46.72} & 426\textbf{\textcolor{darkgreen}{(-88.3\%)}} & \cellcolor{second_best_color} \textbf{47.66} & 540\textbf{\textcolor{darkgreen}{(-91.2\%)}} & \textbf{45.06} & 641\textbf{\textcolor{darkgreen}{(-90.5\%)}} & \cellcolor{second_best_color} \textbf{22.26} & 632\textbf{\textcolor{darkgreen}{(-89.8\%)}} & 43.20 & 592\textbf{\textcolor{darkgreen}{(-87.1\%)}} & \textbf{44.79}\textbf{\textcolor{darkgreen}{(+1.8\%)}} & \textbf{526}\textbf{\textcolor{darkgreen}{(-89.7\%)}} \\
\midrule
AutoRefine$^{\vartriangle}$ & 45.98 & 4,730 & \cellcolor{second_best_color} \textbf{64.73} & 3,903 & 46.84 & 4,109 & 44.79 & 4,962 & 42.95 & 4,879 & 20.52 & 5,614 & \textbf{43.20} & 4,053 & 44.14 & 4,607 \\
\textbf{LatentRAG}$^{\vartriangle}$ & \textbf{47.59} & 368\textbf{\textcolor{darkgreen}{(-92.2\%)}} & 61.54 & 365\textbf{\textcolor{darkgreen}{(-90.7\%)}} & \cellcolor{best_color} \textbf{47.82} & 383\textbf{\textcolor{darkgreen}{(-90.7\%)}} & \cellcolor{best_color} \textbf{48.10} & 467\textbf{\textcolor{darkgreen}{(-90.6\%)}} & \cellcolor{second_best_color} \textbf{45.24} & 539\textbf{\textcolor{darkgreen}{(-89.0\%)}} & \cellcolor{best_color} \textbf{22.92} & 540\textbf{\textcolor{darkgreen}{(-90.4\%)}} & 40.80 & 514\textbf{\textcolor{darkgreen}{(-87.3\%)}} & \cellcolor{second_best_color} \textbf{44.86}\textbf{\textcolor{darkgreen}{(+1.6\%)}} & \textbf{454}\textbf{\textcolor{darkgreen}{(-90.1\%)}} \\
\midrule
\rowcolor{gray!30} \multicolumn{17}{c}{\textbf{harrier-oss-v1-270m}} \\
\midrule
Search-R1$^{\lozenge}$ & 44.40 & 3,510 & \textbf{62.80} & 4,983 & \textbf{47.14} & 3,351 & 44.88 & 6,270 & 44.16 & 6,864 & 18.62 & 6,613 & \textbf{40.00} & 4,944 & \textbf{43.14} & 5,219 \\
\textbf{LatentRAG}$^{\lozenge}$ & \textbf{46.15} & 485\textbf{\textcolor{darkgreen}{(-86.2\%)}} & 60.40 & 444\textbf{\textcolor{darkgreen}{(-91.1\%)}} & 45.41 & 497\textbf{\textcolor{darkgreen}{(-85.2\%)}} & \textbf{47.27} & 614\textbf{\textcolor{darkgreen}{(-90.2\%)}} & \textbf{44.99} & 698\textbf{\textcolor{darkgreen}{(-89.8\%)}} & \textbf{19.90} & 708\textbf{\textcolor{darkgreen}{(-89.3\%)}} & 34.40 & 636\textbf{\textcolor{darkgreen}{(-87.1\%)}} & 42.65\textbf{\textcolor{red}{(-1.1\%)}} & \textbf{583}\textbf{\textcolor{darkgreen}{(-88.8\%)}} \\
\midrule
AutoRefine$^{\vartriangle}$ & 43.80 & 4,457 & \textbf{64.32} & 3,899 & \cellcolor{second_best_color} \textbf{47.68} & 4,153 & 45.40 & 4,703 & 43.31 & 4,910 & 20.15 & 5,511 & \textbf{43.20} & 4,360 & \textbf{43.98} & 4,570 \\
\textbf{LatentRAG}$^{\vartriangle}$ & \textbf{45.68} & 389\textbf{\textcolor{darkgreen}{(-91.3\%)}} & 60.98 & 391\textbf{\textcolor{darkgreen}{(-90.0\%)}} & 45.92 & 409\textbf{\textcolor{darkgreen}{(-90.2\%)}} & \textbf{46.74} & 507\textbf{\textcolor{darkgreen}{(-89.2\%)}} & \textbf{44.58} & 602\textbf{\textcolor{darkgreen}{(-87.7\%)}} & \textbf{20.56} & 579\textbf{\textcolor{darkgreen}{(-89.5\%)}} & 35.20 & 559\textbf{\textcolor{darkgreen}{(-87.2\%)}} & 42.81\textbf{\textcolor{red}{(-2.7\%)}} & \textbf{491}\textbf{\textcolor{darkgreen}{(-89.3\%)}} \\
\midrule
\rowcolor{gray!30} \multicolumn{17}{c}{\textbf{F2LLM-v2-330M}} \\
\midrule
Search-R1$^{\lozenge}$ & 43.35 & 3,717 & \textbf{61.04} & 5,394 & \textbf{44.92} & 3,620 & 41.99 & 6,228 & 41.53 & 6,765 & \textbf{18.54} & 6,690 & \textbf{36.00} & 5,183 & 41.05 & 5,371 \\
\textbf{LatentRAG}$^{\lozenge}$ & \textbf{45.54} & 484\textbf{\textcolor{darkgreen}{(-87.0\%)}} & 58.56 & 432\textbf{\textcolor{darkgreen}{(-92.0\%)}} & 44.21 & 448\textbf{\textcolor{darkgreen}{(-87.6\%)}} & \textbf{44.58} & 593\textbf{\textcolor{darkgreen}{(-90.5\%)}} & \textbf{43.02} & 679\textbf{\textcolor{darkgreen}{(-90.0\%)}} & 18.37 & 667\textbf{\textcolor{darkgreen}{(-90.0\%)}} & \textbf{36.00} & 602\textbf{\textcolor{darkgreen}{(-88.4\%)}} & \textbf{41.47}\textbf{\textcolor{darkgreen}{(+1.0\%)}} & \textbf{558}\textbf{\textcolor{darkgreen}{(-89.6\%)}} \\
\midrule
AutoRefine$^{\vartriangle}$ & 43.82 & 4,483 & \textbf{62.53} & 4,204 & \textbf{45.17} & 4,271 & 42.63 & 4,992 & 41.58 & 5,101 & 19.03 & 5,412 & \textbf{40.80} & 4,381 & \textbf{42.22} & 4,692 \\
\textbf{LatentRAG}$^{\vartriangle}$ & \textbf{45.32} & 403\textbf{\textcolor{darkgreen}{(-91.0\%)}} & 58.83 & 386\textbf{\textcolor{darkgreen}{(-90.8\%)}} & 44.72 & 417\textbf{\textcolor{darkgreen}{(-90.2\%)}} & \textbf{44.74} & 499\textbf{\textcolor{darkgreen}{(-90.0\%)}} & \textbf{44.08} & 583\textbf{\textcolor{darkgreen}{(-88.6\%)}} & \textbf{19.82} & 566\textbf{\textcolor{darkgreen}{(-89.5\%)}} & 36.80 & 552\textbf{\textcolor{darkgreen}{(-87.4\%)}} & 42.04\textbf{\textcolor{red}{(-0.4\%)}} & \textbf{487}\textbf{\textcolor{darkgreen}{(-89.6\%)}} \\
\bottomrule 
\end{tabular}
\end{adjustbox}
\end{table}

\subsection{Latent Decoding}\label{sec:latent_decoding}

To improve transparency of the decision-making process and enhance latent representation learning, we introduce a latent decoding objective. The key idea is to optimize the LLM to reconstruct the corresponding natural language sequences directly from the generated latent tokens. 

We add projector modules $\Projtau$ and $\Projs$ to map latent thought and subquery tokens into the LLM input space, respectively. The projector modules follow the same structure as the projector introduced in Sec~\ref{sec:latent_retrieval}. The projected latent thought tokens or latent subquery tokens are then fed into the LLM to decode the corresponding natural language thought $\tau_t$ or subquery $s_t$:
\begin{gather}
\tau_t = \gLLM( \Projtau (H^{\tau}_{t}); \thetaLLM ), \quad\quad s_t = \gLLM( \Projs (H^{s}_{t}); \thetaLLM ).
\end{gather}
The prompts used to format these inputs are provided in Appendix~\ref{app:prompts}. The decoding process is optimized using the standard cross-entropy loss between the generated sequence and the corresponding natural language target. 
This results in two decoding losses: a thought decoding loss $\setL_{\mathrm{dec}}^{\tau}$ and a subquery decoding loss $\setL_{\mathrm{dec}}^{s}$. The latent decoding loss is the combination of both terms:
\begin{gather}\label{eq:latent_decoding_loss}
\setL_{\mathrm{dec}} = \setL_{\mathrm{dec}}^{\tau} + \setL_{\mathrm{dec}}^{s}.
\end{gather}
During inference, this latent decoding process is optional, allowing the LLM agent to perform reasoning and retrieval entirely in the latent space for efficiency. When required, latent tokens can be decoded into natural language for transparency. Since each decoding process depends only on its corresponding latent tokens, all thoughts and subqueries across multiple steps can be decoded in parallel, thus reducing the latency of generating these natural language sequences.

\subsection{Overall Training Objective}\label{sec:training_procedure}

The overall training objective is defined as a weighted combination of the generation loss, retrieval loss, and latent decoding loss:
\begin{gather}
\setL = \setL_{\mathrm{gen}} + \lambda_{\mathrm{ret}}\setL_{\mathrm{ret}} + \setL_{\mathrm{dec}},
\end{gather}
where $\lambda_{\mathrm{ret}}$ controls the relative scale of the retrieval loss. We do not introduce additional scaling factors for $\setL_{\mathrm{gen}}$ and $\setL_{\mathrm{dec}}$ since both are derived from the standard LLM cross-entropy objective and thus have comparable magnitudes.

\section{Experiments}\label{sec:experiments}

In this section, we conduct experiments to evaluate LatentRAG in terms of effectiveness, efficiency, scalability, and adaptability. Additional experiments and analyses are provided in Appendix~\ref{app:more_experimental_results}.

\subsection{Experimental Setup}

\paragraph{Datasets.} We evaluate LatentRAG using seven common benchmark QA datasets, comprising three general QA datasets (NQ \citep{kwiatkowski2019natural}, TriviaQA \citep{joshi2017triviaqa}, and PopQA \citep{mallen2023not}) and four multi-hop QA datasets (HotpotQA \citep{yang2018hotpotqa}, 2wiki \citep{ho2020constructing}, Musique \citep{trivedi2022musique}, and Bamboogle \citep{press2023measuring}). We use the 2018 Wikipedia dump \citep{karpukhin2020dense} as the corpus for retrieval. More details of the datasets can be found in Appendix~\ref{app:dataset_details}.

\paragraph{Baselines.} We compare LatentRAG against a diverse set of baselines covering direct inference (Direct Infer), traditional single-step RAG (Naive RAG \citep{lewis2020retrieval}), prompt-based agentic RAG (Iter-RetGen \citep{shao2023enhancing}, Search-o1 \citep{li2025search}), and training-based agentic RAG (R1-Searcher \citep{song2025r1}, ZeroSearch \citep{sun2025zerosearch}, DeepRAG \citep{guan2026deeprag}, Search-R1 \citep{jin2025search}, AutoRefine \citep{shi2025search}).

\paragraph{Implementation details.} Following previous works \citep{jin2025search, shi2025search}, we adopt Qwen2.5-7B \citep{qwen2.5} as the default LLM for all methods. For training-based baselines, we utilize their released model weights to ensure the faithful reproduction of their reported performance. Training trajectories are constructed from a combined training set of NQ and HotpotQA using Search-R1 and AutoRefine. Variants trained on trajectories generated by Search-R1 and AutoRefine are denoted as \textbf{LatentRAG$^{\lozenge}$} and \textbf{LatentRAG$^{\vartriangle}$}, respectively. To reduce computational costs, we conduct main experiments using lightweight retrieval models with fewer than 1B parameters, which are among the top-performing models on the MTEB leaderboard \citep{muennighoff2023mteb} and cover diverse model architectures, including Qwen3-Embedding-0.6B \citep{zhang2025qwen3}, e5-base-v2 \citep{wang2022text}, jina-embeddings-v5-text-nano \citep{akram2026jina}, harrier-oss-v1-270m\footnote{\url{https://huggingface.co/microsoft/harrier-oss-v1-270m}}, and F2LLM-v2-330M \citep{zhang2026f2llm}. Unless otherwise specified, we use Qwen3-Embedding-0.6B as the default retriever. To evaluate the trade-off between performance and latency, we report the exact match (EM) score \citep{jin2025search} and the average latency per question. Latency is measured on a single NVIDIA H100 GPU with 94 GB memory by default. More implementation details are in Appendix~\ref{app:implementation_details}.

\subsection{Main Results}

\paragraph{Overall performance and latency.} As shown in Table~\ref{tab:main_results}, advanced agentic RAG methods such as Search-R1 and AutoRefine achieve superior performance over naive single-step RAG, but incur substantially higher latency, with an average overhead of around 15$\times$ the latency in single-step RAG. This latency gap is more pronounced on multi-hop QA datasets. In contrast, LatentRAG trained on trajectories from Search-R1 and AutoRefine achieves comparable performance, with relative differences within 5\%, while significantly reducing latency by approximately 90\%. This advantage holds consistently across diverse retrieval models. Fig.~\ref{fig:performance_stage_inference_time} shows that LatentRAG significantly reduces latency in thought and subquery generation.

Compared to other retrieval models, we observe a relatively large performance drop when using e5-base-v2. To investigate the source of this discrepancy, we analyze the embedding spaces of different retrieval models. As shown in Fig.~\ref{fig:cosine_angle_distribution} in the Appendix, e5-base-v2 exhibits severe anisotropy \citep{li2020sentence, zhou2021isobn}, indicating that the embeddings produced by the model are highly concentrated within a narrow cone on a hypersphere. This skewed distribution makes it difficult for the LLM to adapt to the retrieval space. More analysis is provided in Appendix~\ref{app:retriever_embedding_space_analysis}.

\begin{wraptable}{r}{0.5\textwidth}
\vspace{-12pt}
\small{\caption{\textbf{Latency with and without decoding.}}\label{tab:latency_with_decoding}}
\vspace{0pt}
\centering
\begin{adjustbox}{max width = 1.0\linewidth}
\begin{tabular}{l|cc}
\toprule
\textbf{Methods} & \textbf{Lat. (ms)}$^{\downarrow}$ & \textbf{Max Len. Ratio(\%)} \\
\midrule
Search-R1$^{\lozenge}$ & 5,372 & 37.74 \\
\textbf{LatentRAG}$^{\lozenge}$ w/o decoding & 593\textbf{\textcolor{darkgreen}{(-89.0\%)}} & -- \\
\textbf{LatentRAG}$^{\lozenge}$ w/ decoding & 1,970\textbf{\textcolor{darkgreen}{(-63.3\%)}} & 33.70 \\
\midrule
AutoRefine$^{\vartriangle}$ & 4,827 & 43.14 \\
\textbf{LatentRAG}$^{\vartriangle}$ w/o decoding & 512\textbf{\textcolor{darkgreen}{(-89.4\%)}} & -- \\
\textbf{LatentRAG}$^{\vartriangle}$ w/ decoding & 2,540\textbf{\textcolor{darkgreen}{(-47.4\%)}} & 42.13 \\
\bottomrule 
\end{tabular}
\end{adjustbox}
\vspace{-6pt}
\end{wraptable}

\paragraph{Latent decoding efficiency.} Latent decoding is an option for improving transparency at the cost of additional latency. To quantify this overhead, Table~\ref{tab:latency_with_decoding} reports the latency of LatentRAG with and without latent decoding. Latent decoding increases the overall latency of LatentRAG by approximately 4–5$\times$. Nevertheless, it still reduces latency by 63.3\% and 47.4\% compared to Search-R1 and AutoRefine, respectively. The efficiency gain stems from the removal of sequential dependencies, enabling parallel decoding across steps. The actual speedup is bounded by the longest sequence in the batch, which determines the number of decoding steps required. We report the \textit{max length ratio} in Table~\ref{tab:latency_with_decoding}, defined as the fraction of tokens in the longest thought or subquery sequence over the total decoding length. A higher ratio indicates a more imbalanced distribution of sequence lengths. In particular, LatentRAG$^{\vartriangle}$ exhibits a larger max length ratio, which explains its less pronounced efficiency gains. Further analysis is provided in Appendix~\ref{app:latent_decoding_efficiency_analysis}, along with case studies of decoded examples in Appendix~\ref{app:case_studies}.

\begin{figure*}[t]
\centering
\includegraphics[width=1\linewidth]{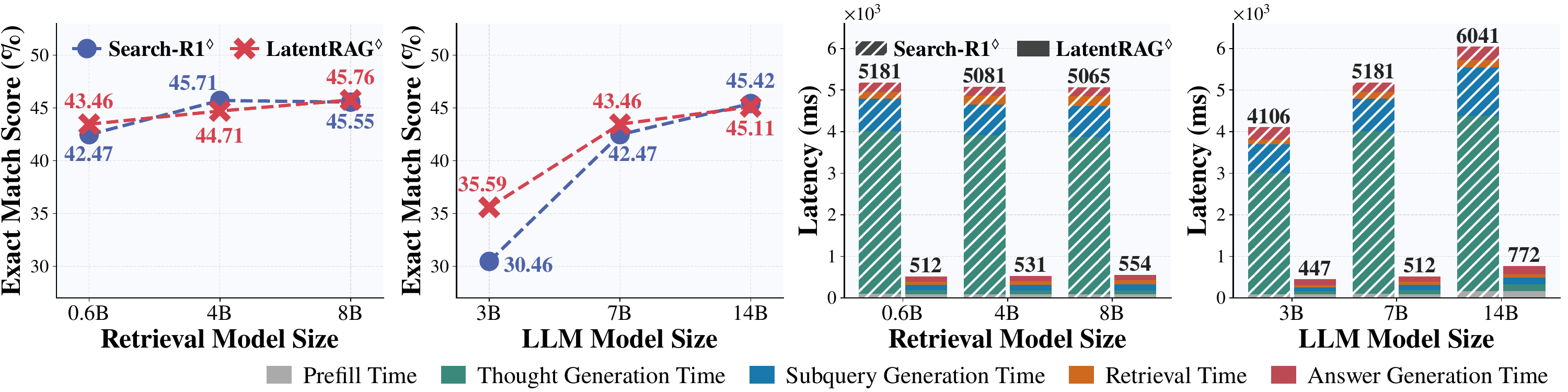}
\caption{\textbf{Performance and latency results across different retrieval model and LLM sizes.}}
\label{fig:model_size_scaling}
\end{figure*}

\paragraph{Scaling model size.} We study scalability along two orthogonal dimensions. For retrieval model scaling, we evaluate Qwen3-Embedding-0.6B, 4B, and 8B \citep{zhang2025qwen3} with a fixed 7B LLM. For LLM scaling, we evaluate Qwen2.5-3B, 7B, and 14B with a fixed Qwen3-Embedding-0.6B retrieval model. Larger retrieval models produce higher-dimensional embeddings, resulting in a substantially larger index that cannot fit on a single GPU. To ensure a fair comparison across different model sizes, we use three H100 GPUs for retrieval deployment and one for the LLM across all scaling experiments.

As shown in Fig.~\ref{fig:model_size_scaling}, performance improves with increasing model size along both dimensions. Scaling the retrieval model introduces negligible latency overhead, as the retrieval process can be efficiently parallelized. In contrast, scaling the LLM leads to substantial latency increases for SearchR1 due to increased decoding time. Our method achieves comparable performance across all settings and yields improvements in the 3B LLM setting while significantly reducing inference latency.

\paragraph{Cross-retriever adaptation.} Since LatentRAG requires aligning the representation spaces of the LLM and retrieval model, switching to a different retrieval model requires adaptation. We investigate whether LatentRAG can be adapted to a new retrieval model without finetuning the LLM or the retrieval model. Specifically, we adapt LatentRAG trained with Qwen3-Embedding-0.6B (Qwen3) to jina-embeddings-v5-text-nano (Jina) by finetuning only the retrieval projector. As shown in Table~\ref{tab:cross_retriever_results}, the adapted model achieves performance similar to that of LatentRAG trained from scratch with Jina. Both variants achieve a slight relative improvement over the Search-R1 baseline. These results demonstrate that finetuning only the retrieval projector is sufficient for cross-retriever adaptation.

\begin{table}[t]
\small{\caption{\textbf{Cross-retriever adaptation.} This table reports exact match scores (\%). LatentRAG adapted from Qwen3 to Jina, finetuning only the retrieval projector, achieves performance comparable to that of the variant trained with the Jina retrieval model from scratch.}
\label{tab:cross_retriever_results}} \centering
\setlength{\tabcolsep}{6pt}
\begin{adjustbox}{max width = 1.0\linewidth}
\begin{tabular}{l|ccccccc|c}
\toprule
\textbf{Methods} & \textbf{NQ} & \textbf{TriviaQA} & \textbf{PopQA} & \textbf{HotpotQA} & \textbf{2wiki} & \textbf{Musique} & \textbf{Bamboogle} & \textbf{Average} \\
\midrule
\textbf{Search-R1$^{\lozenge}$} (Jina) & 45.79 & \textbf{63.16} & 46.54 & 44.31 & 42.95 & 20.40 & \textbf{44.80} & 43.99 \\
\textbf{LatentRAG$^{\lozenge}$} (Jina) & 47.37 & 61.25 & 46.72 & \textbf{47.66} & \textbf{45.06} & \textbf{22.26} & 43.20 & \textbf{44.79} \textbf{\textcolor{darkgreen}{(+1.8\%)}} \\
\textbf{LatentRAG$^{\lozenge}$} (Qwen3$\rightarrow$Jina) & \textbf{47.62} & 61.25 & \textbf{46.89} & \textbf{47.66} & 44.51 & 21.47 & 41.60 & 44.43 \textbf{\textcolor{darkgreen}{(+1.0\%)}} \\
\bottomrule
\end{tabular}
\end{adjustbox}
\end{table}

We further investigate the data efficiency of cross-retriever adaptation. Specifically, we examine whether adaptation can be achieved with limited training data or even using only synthetic data. We first randomly sample 5,000 queries from the original training set to construct adaptation trajectories, corresponding to less than 10\% of the original training data, and compare the adapted model with a model trained from scratch using the same amount of data. We then consider a more challenging setting where adaptation uses only 5,000 synthetic queries, with no real query data. The synthetic queries are generated by the base model Qwen2.5-7B from randomly sampled documents in the Wikipedia corpus. The prompt used for query generation is provided in Appendix~\ref{app:prompts}. Since each query is generated from a single document, these synthetic queries are generally simple and differ substantially from the complex multi-hop queries in the evaluation datasets.

\begin{table}[t]
\small{\caption{\textbf{Data efficiency of cross-retriever adaptation.} This table reports exact match scores (\%). When using a limited amount of training data or even only synthetic data, LatentRAG adapted from Qwen3 to Jina achieves significantly better performance than the model trained from scratch.}
\label{tab:data_efficiency_of_cross_retriever_adaptation}} \centering
\setlength{\tabcolsep}{4pt}
\begin{adjustbox}{max width = 1.0\linewidth}
\begin{tabular}{l|ccccccc|c}
\toprule
\textbf{Methods} & \textbf{NQ} & \textbf{TriviaQA} & \textbf{PopQA} & \textbf{HotpotQA} & \textbf{2wiki} & \textbf{Musique} & \textbf{Bamboogle} & \textbf{Average} \\
\midrule
\textbf{LatentRAG$^{\lozenge}$} (Jina, full data) & 47.37 & \textbf{61.25} & 46.72 & \textbf{47.66} & \textbf{45.06} & \textbf{22.26} & \textbf{43.20} & \textbf{44.79} \\
\textbf{LatentRAG$^{\lozenge}$} (Qwen3$\rightarrow$Jina, full data) & \textbf{47.62} & \textbf{61.25} & \textbf{46.89} & \textbf{47.66} & 44.51 & 21.47 & 41.60 & 44.43 \\
\midrule
\textbf{LatentRAG$^{\lozenge}$} (Jina, \textbf{5,000 real data}) & 35.12 & 47.84 & 34.61 & 33.99 & 26.79 & 10.67 & 22.40 & 30.20 \\
\textbf{LatentRAG$^{\lozenge}$} (Qwen3$\rightarrow$Jina, \textbf{5,000 real data}) & \textbf{46.93} & \textbf{58.85} & \textbf{45.29} & \textbf{46.06} & \textbf{43.21} & \textbf{19.36} & \textbf{40.80} & \textbf{42.93} \\
\midrule
\textbf{LatentRAG$^{\lozenge}$} (Jina, \textbf{5,000 synthetic data}) & 25.18 & 40.13 & 30.07 & 20.65 & 15.93 & 3.56 & 7.20 & 20.39 \\
\textbf{LatentRAG$^{\lozenge}$} (Qwen3$\rightarrow$Jina, \textbf{5,000 synthetic data}) & \textbf{45.12} & \textbf{57.62} & \textbf{44.66} & \textbf{43.92} & \textbf{43.71} & \textbf{17.63} & \textbf{39.20} & \textbf{41.69} \\
\bottomrule
\end{tabular}
\end{adjustbox}
\end{table}

As shown in Table~\ref{tab:data_efficiency_of_cross_retriever_adaptation}, using only 5,000 training trajectories, LatentRAG trained from scratch exhibits substantially worse performance than the variant trained using the full data, suggesting that 5,000 training trajectories are insufficient to learn effective retrieval policies from scratch. In contrast, LatentRAG adapted from Qwen3 to Jina achieves performance comparable to that of the full-data variant, demonstrating that cross-retriever adaptation can be achieved with relatively limited training data compared to training from scratch. More importantly, even when using only 5,000 synthetic queries, the adapted model substantially outperforms the model trained from scratch, demonstrating that cross-retriever adaptation remains effective even when real training data are unavailable.

\subsection{Ablation Studies}

\begin{wraptable}{r}{0.5\textwidth}
\vspace{-12pt}
\small{\caption{\textbf{Ablation study results.} Overlap indicates the proportion of documents retrieved by Search-R1 that are also retrieved by each model.}\label{tab:ablation}}
\vspace{0pt}
\centering
\begin{adjustbox}{max width = 1.0\linewidth}
\begin{tabular}{l|ccc}
\toprule
\textbf{Methods} & \textbf{EM(\%)}$^{\uparrow}$ & \textbf{Success(\%)}$^{\uparrow}$ & \textbf{Overlap(\%)} \\
\midrule
Search-R1$^{\lozenge}$ & 42.47 & 60.15 & 100.00 \\
\midrule
\rowcolor{gray!30} \textbf{LatentRAG}$^{\lozenge}$(w/ KL loss) & \textbf{43.46} & \textbf{61.27} & 59.41 \\
- (i) Cosine loss & 42.55 & 60.76 & 68.31 \\
- (ii) InfoNCE loss & 41.86 & 58.60 & 47.08 \\
- (iii) w/o retriever & 41.85 & 59.07 & 50.92\\
- (iv) w/o decoding & 40.61 & 60.64 & 57.38 \\
\bottomrule 
\end{tabular}
\end{adjustbox}
\vspace{-6pt}
\end{wraptable}

We conduct ablation studies on key design choices to validate their effectiveness. Specifically, we replace the KL-based retrieval objective in Eq.~\ref{eq:retrieval_loss} with two alternative choices: (i) a cosine loss, which directly minimizes the cosine distance between the latent subquery embedding $\vv_{s^{\ell}_{t}}$ and the corresponding reference subquery embedding $\vv_{s_t}'$, and (ii) a standard InfoNCE loss \citep{oord2018representation}, which is widely used for training retrieval models. We further consider two ablation settings: (iii) removing the pretrained retrieval model and relying solely on the LLM to produce subquery embeddings, and (iv) removing the latent decoding loss in Eq.~\ref{eq:latent_decoding_loss} during training. We report the average EM score as well as two retrieval related metrics: (a) retrieval success rate, defined as the proportion of successful iterative retrievals where the retrieved documents contain the ground truth answer, and (b) retrieval overlap, defined as the proportion of documents retrieved by Search-R1 that are also retrieved by our model.

As shown in Table~\ref{tab:ablation}, LatentRAG with the proposed KL-based objective achieves better EM score and retrieval success rate compared to the cosine and InfoNCE alternatives. The cosine loss yields the highest retrieval overlap ratio, indicating closer imitation of the teacher model Search-R1. However, its performance is lower than that of the KL-based variant, suggesting that overly aligning with the teacher model may limit model capacity and lead to suboptimal performance. Removing the pretrained retrieval model also degrades the performance, highlighting the importance of the inductive bias provided by the pretrained retrieval model. Finally, removing the latent decoding loss leads to performance degradation, suggesting that latent decoding not only improves transparency at inference time, but also facilitates the learning of latent representations during training.

\section{Conclusion}\label{sec:conclusion}

In this paper, we propose LatentRAG, an efficient agentic RAG framework that shifts both reasoning and retrieval from discrete language space to continuous latent space. Experiments show that Latent\-RAG achieves performance comparable to existing agentic RAG methods while reducing latency by approximately 90\%. To improve transparency, the latent tokens can be optionally decoded into natural language with additional latency overhead, while still achieving an overall 40–60\% reduction in latency compared to the baselines. Experiments across different model scales demonstrate the general applicability of LatentRAG. Cross-retriever experiments further demonstrate the adaptability of LatentRAG to new retrieval models, recovering approximately 95\% of the performance by finetuning only the retrieval projector with a small amount of training data or even only synthetic data. Additional experiments in Appendix~\ref{app:more_experimental_results} further demonstrate the effectiveness and efficiency of LatentRAG under different settings, including its superior performance on long-horizon datasets and its latency advantages over baseline methods with vLLM optimization and speculative decoding.

\begin{ack}

This work is supported by the AI4Intelligence project with file number KICH1.VE01.20.011, partly financed by the Dutch Research Council (Nederlandse Organisatie voor Wetenschappelijk Onderzoek, NWO). This work used the Dutch national e-infrastructure with the support of the SURF Cooperative using grant number EINF-15515. The authors declare no competing interests.

\end{ack}

{
\small
\bibliography{main}
\bibliographystyle{unsrtnat}
}


\newpage
\clearpage
\appendix

\section{Broader Impacts and Limitations}\label{app:broader_impacts_and_limitations}

\paragraph{Broader impacts.} This work proposes an efficient agentic RAG framework that performs both reasoning and retrieval in the latent space. The proposed approach could be applied to a wide range of information retrieval scenarios, such as legal or clinical question answering \citep{lai2024large, singhal2025toward}, and improve overall efficiency in these tasks. More broadly, this work raises a fundamental question: \textbf{Should information retrieval be redesigned for AI agents?} Humans naturally use words to search for information, and traditional retrieval systems therefore typically rely on text queries. However, as retrieval systems are increasingly used by AI agents rather than directly by humans, it is worth reconsidering this text-centric paradigm. Both AI agents and retrieval systems operate on continuous latent representations, making it possible for AI agents to search without first translating their internal latent representations into text. This opens up new opportunities to rethink information retrieval in the era of agentic AI and develop retrieval systems better suited to AI agents.

\paragraph{Limitations and future work.} Our method relies on supervised finetuning over trajectories generated by existing agentic RAG methods, and its performance is therefore partly bounded by the quality of the training data. This hinders the model from directly learning an optimal retrieval policy through interactions with the retrieval system. Nevertheless, our approach yields strong and efficient initial models that serve as an effective foundation for future research. Future work could investigate reinforcement learning to further improve the SFT-trained model by encouraging exploration and exploitation. Additionally, while LatentRAG takes a first step toward reducing agentic RAG latency through latent reasoning and retrieval, it has not yet been integrated with existing inference optimization frameworks. Future work could explore adapting frameworks such as vLLM to LatentRAG to enable efficient batched inference and improve overall throughput in practical deployments.

\section{Implementation Details}\label{app:implementation_details}

\paragraph{Training data construction.} As described in the main paper, we combine the training sets from NQ and HotpotQA to construct a unified training dataset. We then build training trajectories using interaction data generated by Search-R1 and AutoRefine on this unified training dataset. Each trajectory consists of the question, intermediate reasoning thoughts, subqueries, retrieved document chunks, and the final generated answer. AutoRefine introduces an additional refinement stage to improve the initially retrieved documents. To maintain a consistent trajectory format with Search-R1, we merge the refinement text into the reasoning thoughts. We retain only those trajectories that produce a correct final answer for training. To facilitate finer-grained control over different components in the generated trajectories, we introduce a set of special tokens to explicitly mark structural elements in the output, such as $\tagtoken{Answer}\ldots\tagtoken{\texttt{/}Answer}$. In contrast, these tags are typically tokenized into multiple subword units in Search-R1 and AutoRefine. This difference may introduce minor variations in generation time. However, its impact is negligible compared to the overall latency reduction achieved by our framework.

\paragraph{Computing resources \& parallelization strategies.} For training, we optimize LatentRAG on a single compute node equipped with two NVIDIA H100 GPUs, each with 94 GB of memory. Each training job takes about 24 to 48 hours to complete. To reduce GPU memory consumption, we enable gradient checkpointing to minimize the storage of intermediate activations. For distributed training, we adopt DeepSpeed ZeRO1\footnote{\url{https://github.com/deepspeedai/DeepSpeed}}, which shards the optimizer states across GPU devices while keeping gradients and model parameters fully replicated. This design avoids additional communication overhead associated with parameter and gradient sharding, thereby maintaining efficient data-parallel training. To address the imbalance in trajectory lengths, we implement a binned batching strategy. Specifically, we partition trajectories into 200 bins according to their lengths and construct each batch by sampling from a single bin. This binned batching strategy ensures that samples within each batch have similar sequence lengths and therefore reduces padding overhead and improves computational efficiency. We use bfloat16 precision and FlashAttention-2\footnote{\url{https://github.com/dao-ailab/flash-attention}} during training. We adopt LoRA\footnote{\url{https://github.com/microsoft/LoRA}} with a rank of 16 for parameter-efficient finetuning, which significantly reduces the number of trainable parameters, thereby lowering memory and computational costs.

For evaluation, we conduct experiments on a single NVIDIA H100 GPU with 94 GB of memory by default. We deploy the retrieval system using Faiss\footnote{\url{https://github.com/facebookresearch/faiss}} on the GPU with half-precision indexing and load the LLM on the same device. To ensure a fair comparison across different methods, we measure both LLM prefill and decoding latency using the standard forward pass implemented in Hugging Face Transformers\footnote{\url{https://huggingface.co/docs/transformers}}. For scaling experiments, larger retrieval models produce higher-dimensional document embeddings that exceed the memory capacity of a single GPU. For example, the index built from Qwen3-Embedding-8B occupies approximately 160 GB even with float16 precision. To accommodate this, for all the scaling experiments, we deploy the retrieval system across three H100 GPUs, while using a separate H100 GPU to serve the LLM. This setup ensures sufficient GPU resources for both retrieval and generation, allowing us to report latency under sufficient computational resources, where system bottlenecks are minimized.

\paragraph{Hyperparameters.} We finetune the model using LoRA with rank 16 and scaling factor 64, applied to all projection weights. The model is trained for 5 epochs with a learning rate of $1\times10^{-4}$. The maximum trajectory length is capped at 3000 tokens. For the KL divergence loss, we set the target distribution based on the similarity scores between queries and documents. Specifically, we select the temperature factor that makes the cumulative probability of the top-3 retrieved documents approach 0.5. In practice, this corresponds to setting the temperature to $\beta = 0.03$ in most cases, while the detailed setting for each case can be found in our \href{https://github.com/zhengyijia/LatentRAG}{source code}. The loss weight for the retrieval objective is set to $\lambda_{\mathrm{ret}} = 1$. For the retrieval model, we remove dropout to reduce noise in the target distribution, while for the LLM we apply a dropout rate of 0.1. We use $m=4$ thought tokens for each thought generation step and $n=16$ subquery tokens for each subquery generation step. The training batch size is set to 16. The model is optimized using the AdamW optimizer with $\beta_1 = 0.9$, $\beta_2 = 0.999$, and a weight decay of 0.01. For retrieval loss calculation, we retrieve the top-16 documents as pseudo-relevant documents and combine them with in-batch negative samples, \ie, the pseudo-relevant documents from other subqueries within the same batch, to form a candidate document set, which will be used to compute the document probability distribution.

\paragraph{Evaluation metrics \& measurement protocol.} We adopt exact match (EM) as the primary performance evaluation metric. The EM score measures whether the final predicted answer exactly matches the ground-truth answer. Before evaluation, both predicted and ground-truth answers are normalized by removing articles (e.g., a, an, the), stripping whitespace, removing punctuation, and converting all text to lowercase. For all retrieval-based methods, we retrieve the top-3 documents per query. The maximum number of retrieval iterations is set to 4. For efficiency, we report latency, which captures the end-to-end response time from receiving a query to generating the final answer. We sample the first 100 questions from each dataset to estimate latency. To enable fine-grained latency analysis, we report a breakdown of the latency across different stages, including prefill, thought generation, subquery generation, retrieval, and answer generation.

Following prior work \citep{jin2025search, shi2025search}, we use Qwen2.5 Instruct for inference-based methods due to its stronger instruction-following capabilities. For training-based baselines, we adopt checkpoints released in the original papers that are based on the Base variant of Qwen2.5, which demonstrated better performance in prior work compared to the Instruct variant under training-based settings \citep{jin2025search}. We also initialize and finetune our model from Qwen2.5 Base for a fair comparison. 

For stage-wise latency measurement, the embedding time of a natural language subquery is included in the retrieval stage. For our method, to reduce the number of vectors transmitted to the retrieval system, we generate subquery embeddings on the model side from the latent tokens and pass only the resulting embedding vector to the retrieval system. Therefore, the embedding computation time is attributed to the subquery generation stage instead of the retrieval stage. This design leads to higher measured subquery generation time and lower retrieval time for our method. However, this difference does not affect the computation of the overall latency.

\section{Prompt Templates}\label{app:prompts}

In this section, we provide all prompt templates used in our framework. Double curly braces \{\{ \texttt{$\cdots$} \}\} denote runtime placeholders. Prompt~\ref{prompt:question_prompt} presents the template for latent thought and subquery generation. The latent thought and subquery tokens are derived from the hidden states at the positions of the corresponding special tokens. An action token is predicted based on the final thought token. If the action token is $\tagtoken{answer}$, the special subquery tokens in the prompt template are replaced with the answer token $\tagtoken{answer}$ to trigger the answer generation process. Prompt~\ref{prompt:thought_decoding_prompt} and Prompt~\ref{prompt:subquery_decoding_prompt} present the templates for latent thought and subquery decoding, respectively. Prompt~\ref{prompt:query_synthetic_prompt} presents the template used to generate synthetic questions from Wikipedia passages. These synthetic questions are used in the cross-retriever adaptation experiments without using real query data.

\begin{promptbox}[label=prompt:question_prompt,width=\linewidth]
{Prompt template for thought and subquery generation}
Answer the following question by reasoning step by step and retrieving necessary information at each step: \\
\{\{\texttt{QUESTION}\}\}\\
$\tagtoken[1]{think} \ldots \tagtoken[m]{think}$$\tagtoken[1]{query} \ldots \tagtoken[n]{query}$\\
$\tagtoken{information}$\\
Doc 1 \{\{\texttt{TOP-$\mathtt{1}$\_DOCUMENT}\}\}\\
Doc 2 \{\{\texttt{TOP-$\mathtt{2}$\_DOCUMENT}\}\}\\
Doc 3 \{\{\texttt{TOP-$\mathtt{3}$\_DOCUMENT}\}\}\\
$\tagtoken{\texttt{/}information}$\\
$\cdots$\\
$\tagtoken[1]{think} \ldots \tagtoken[m]{think}$$\tagtoken[1]{query} \ldots \tagtoken[n]{query}$\\
\end{promptbox}

\begin{promptbox}[label=prompt:thought_decoding_prompt,width=\linewidth]
{Prompt template for latent thought decoding}
Decode the thought based on the latent representation: \{\{\texttt{LATENT\_THOUGHT\_TOKENS}\}\}
\end{promptbox}

\begin{promptbox}[label=prompt:subquery_decoding_prompt,width=\linewidth]
{Prompt template for latent subquery decoding}
Decode the subquery based on the latent representation: \{\{\texttt{LATENT\_SUBQUERY\_TOKENS}\}\}
\end{promptbox}

\begin{promptbox}[label=prompt:query_synthetic_prompt,width=\linewidth]
{Prompt template for generating synthetic queries}
Generate a question answer pair based on the given passage. The answer should be short (e.g., a single entity, a short phrase, or "yes"/"no"). Respond with a JSON object that contains a question and an answer, following the JSON schema below:\\\\
\textasciigrave\textasciigrave\textasciigrave
\begin{verbatim}
{
    "type": "object",
    "properties": {
        "question": {
            "type": "string",
        }, 
        "answer": {
            "type": "string",
        }
    },
    "required": ["question", "answer"]
}
\end{verbatim}
\textasciigrave\textasciigrave\textasciigrave\\
Passage:\\\\
\{\{\texttt{PASSAGE}\}\}
\end{promptbox}

\section{Dataset Description}\label{app:dataset_details}

We conduct our experiments on seven benchmark QA datasets, following previous works \citep{jin2025search, shi2025search}. These datasets include three general QA datasets (NQ \citep{kwiatkowski2019natural}, TriviaQA \citep{joshi2017triviaqa}, and PopQA \citep{mallen2023not}) and four multi-hop QA datasets (HotpotQA \citep{yang2018hotpotqa}, 2wiki \citep{ho2020constructing}, Musique \citep{trivedi2022musique}, and Bamboogle \citep{press2023measuring}). Instead of using the original documents provided by each dataset as the retrieval corpus, we follow \citep{jin2025flashrag} and adopt a more challenging and realistic setting by using the full Wikipedia 2018 dump \citep{karpukhin2020dense} as the corpus. The corpus contains 21,015,324 chunked documents, making retrieval significantly more difficult due to its large scale and diverse content. For training, we use the dataset splits provided by FlashRAG\footnote{\url{https://huggingface.co/datasets/RUC-NLPIR/FlashRAG_datasets}} and train our models on the training sets of NQ and HotpotQA. We evaluate all methods on the test sets (or development sets when test sets are unavailable) of the seven benchmark datasets. Table~\ref{tab:dataset_summary} summarizes the number of QA pairs in each dataset. Bamboogle is a very small dataset containing only 125 samples, which may lead to higher evaluation variance and less stable performance estimates compared to the other benchmark datasets.

\begin{table}[t]
\small{\caption{\textbf{Summary of datasets.}}\label{tab:dataset_summary}}
\centering
\setlength{\tabcolsep}{3pt}
\begin{adjustbox}{max width = 1.0\linewidth}
\begin{tabular}{lccccccc}
\toprule
& \textbf{NQ} & \textbf{TriviaQA} & \textbf{PopQA} & \textbf{HotpotQA} & \textbf{2wiki} & \textbf{Musique} & \textbf{Bamboogle} \\
\midrule
\# Train & 79,168 & -- & -- & 90,447 & -- & -- & -- \\
\# Test & 3,610 & 11,313 & 14,267 & 7,405 & 12,576 & 2,417 & 125 \\
\bottomrule 
\end{tabular}
\end{adjustbox}
\end{table}

\section{More Experimental Results}\label{app:more_experimental_results}

\subsection{Embedding Space Analysis of Retrieval Models}\label{app:retriever_embedding_space_analysis}

In Table~\ref{tab:main_results} of the main paper, compared to other retrieval models, our method exhibits a relatively larger performance drop when using e5-base-v2. To further investigate the source of this discrepancy, we analyze the differences in the geometric properties of the embedding space across different retrieval models. Specifically, for each retrieval model, we generate $\ell_2$-normalized embeddings for the entire Wikipedia corpus. We then compute the mean direction of all document embeddings produced by that model. Next, we measure the cosine similarity and the angular distance between this mean direction and each document embedding and visualize their respective distributions. Any of these distributions skewed toward higher cosine similarities (or lower angles) indicates that the embeddings are concentrated around the mean direction rather than being uniformly distributed over the hypersphere, thus reflecting a stronger anisotropy \citep{li2020sentence, zhou2021isobn} in the embedding space.

\begin{figure*}[t]
\begin{center}
\includegraphics[trim={0cm 0cm 0cm 0cm},clip,width=1.0\linewidth]{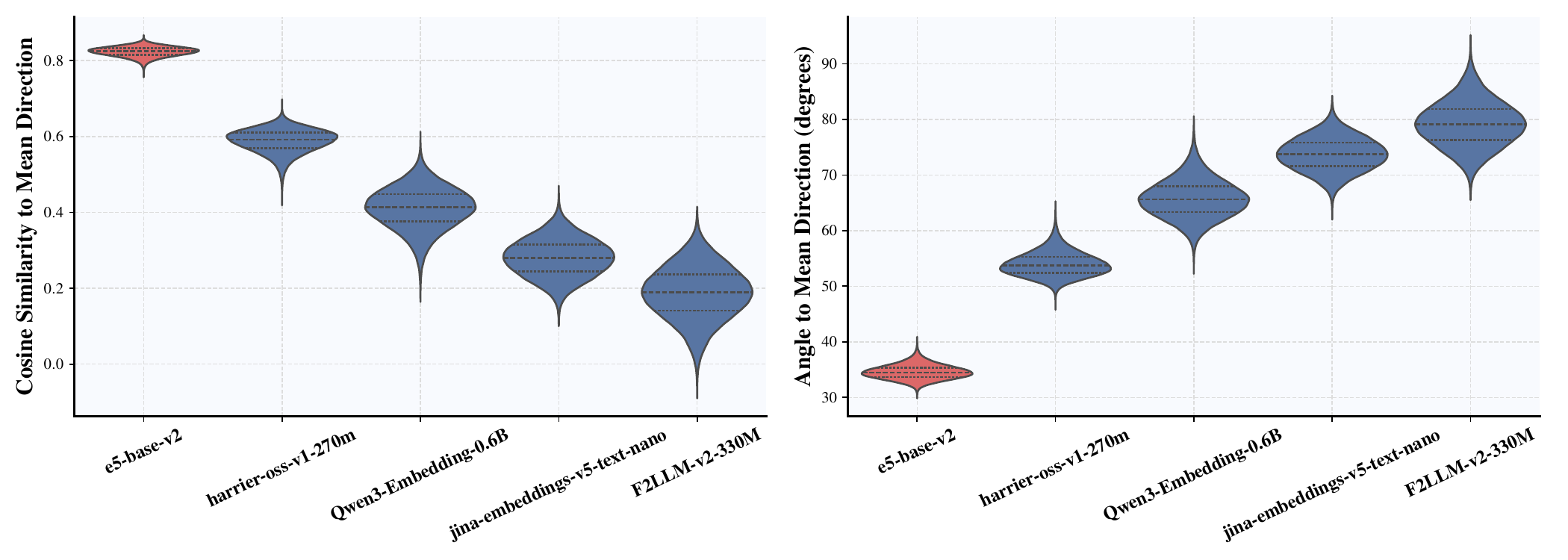}
\end{center}
\caption{\textbf{Distribution of cosine similarity and angle between document embeddings and their mean direction.} We visualize distributions using violin plots. In contrast to other retrieval models, e5-base-v2 yields embeddings with extremely high cosine similarity and small angular deviation, indicating collapse into a narrow cone of the hypersphere and severe anisotropy.}
\label{fig:cosine_angle_distribution}
\end{figure*}

As shown in Fig.~\ref{fig:cosine_angle_distribution}, the embeddings generated by e5-base-v2 exhibit extremely high cosine similarity and low angular deviation with respect to the mean direction, demonstrating severe anisotropy. This suggests that the embeddings are highly concentrated around a narrow region of the hypersphere, rather than being well spread out. As a result, small approximation errors in the embedding space may lead to completely different retrieval outputs, making it challenging to train a model to faithfully approximate the behavior of the original retrieval model. Moreover, such a skewed distribution may force the LLM to deviate from its original representation geometry to adapt to this skewed concentrated space, which could negatively affect the performance of the LLM.

\subsection{Latent Decoding Efficiency Analysis}\label{app:latent_decoding_efficiency_analysis}

\begin{table}[t]
\small{\caption{\textbf{Average latency (ms) with and without latent decoding across all datasets.}}\label{tab:detailed_latent_decoding_latency}}
\centering
\setlength{\tabcolsep}{3pt}
\begin{adjustbox}{max width = 1.0\linewidth}
\begin{tabular}{l|ccccccc|c}
\toprule
\textbf{Methods} & \textbf{NQ} & \textbf{TriviaQA} & \textbf{PopQA} & \textbf{HotpotQA} & \textbf{2wiki} & \textbf{Musique} & \textbf{Bamboogle} & \textbf{Average} \\
\midrule
Search-R1$^{\lozenge}$ & 3,553 & 5,198 & 3,588 & 6,589 & 6,925 & 6,846 & 4,904 & 5,372 \\
\textbf{LatentRAG}$^{\lozenge}$ w/o decoding & 491\textbf{\textcolor{darkgreen}{(-86.2\%)}} & 478\textbf{\textcolor{darkgreen}{(-90.8\%)}} & 501\textbf{\textcolor{darkgreen}{(-86.0\%)}} & 626\textbf{\textcolor{darkgreen}{(-90.5\%)}} & 704\textbf{\textcolor{darkgreen}{(-89.8\%)}} & 730\textbf{\textcolor{darkgreen}{(-89.3\%)}} & 623\textbf{\textcolor{darkgreen}{(-87.3\%)}} & 593\textbf{\textcolor{darkgreen}{(-89.0\%)}} \\
\textbf{LatentRAG}$^{\lozenge}$ w/ decoding & 1,654\textbf{\textcolor{darkgreen}{(-53.5\%)}} & 1,855\textbf{\textcolor{darkgreen}{(-64.3\%)}} & 1,634\textbf{\textcolor{darkgreen}{(-54.5\%)}} & 2,110\textbf{\textcolor{darkgreen}{(-68.0\%)}} & 2,248\textbf{\textcolor{darkgreen}{(-67.5\%)}} & 2,285\textbf{\textcolor{darkgreen}{(-66.6\%)}} & 2,002\textbf{\textcolor{darkgreen}{(-59.2\%)}} & 1,970\textbf{\textcolor{darkgreen}{(-63.3\%)}} \\
\midrule
AutoRefine$^{\vartriangle}$ & 4,782 & 4,223 & 4,397 & 5,344 & 5,264 & 5,553 & 4,224 & 4,827 \\
\textbf{LatentRAG}$^{\vartriangle}$ w/o decoding & 409\textbf{\textcolor{darkgreen}{(-91.4\%)}} & 400\textbf{\textcolor{darkgreen}{(-90.5\%)}} & 422\textbf{\textcolor{darkgreen}{(-90.4\%)}} & 528\textbf{\textcolor{darkgreen}{(-90.1\%)}} & 607\textbf{\textcolor{darkgreen}{(-88.5\%)}} & 639\textbf{\textcolor{darkgreen}{(-88.5\%)}} & 581\textbf{\textcolor{darkgreen}{(-86.2\%)}} & 512\textbf{\textcolor{darkgreen}{(-89.4\%)}} \\
\textbf{LatentRAG}$^{\vartriangle}$ w/ decoding & 3,706\textbf{\textcolor{darkgreen}{(-22.5\%)}} & 2,403\textbf{\textcolor{darkgreen}{(-43.1\%)}} & 2,576\textbf{\textcolor{darkgreen}{(-41.4\%)}} & 2,571\textbf{\textcolor{darkgreen}{(-51.9\%)}} & 2,102\textbf{\textcolor{darkgreen}{(-60.1\%)}} & 2,300\textbf{\textcolor{darkgreen}{(-58.6\%)}} & 2,125\textbf{\textcolor{darkgreen}{(-49.7\%)}} & 2,540\textbf{\textcolor{darkgreen}{(-47.4\%)}} \\
\bottomrule 
\end{tabular}
\end{adjustbox}
\end{table}

As discussed in the main paper, latent decoding improves transparency at the cost of additional latency. A good property of our method is that the decoding of thoughts and subqueries is conditionally independent given the latent tokens. This property allows parallel decoding across different steps, which is more efficient than the sequential generation in existing agentic RAG methods.

To quantify the effect of reduced latency enabled by our parallel decoding strategy, we report detailed latency measurements across multiple datasets and compare them with baseline methods. As shown in Table~\ref{tab:detailed_latent_decoding_latency}, using latent decoding increases latency by approximately 4–5$\times$ compared to the setting without latent decoding. Nevertheless, compared to corresponding baseline methods, our method with latent decoding reduces overall latency by approximately 23-68\% across different datasets. 

\begin{wrapfigure}{r}{0.4\textwidth}
\vspace{-12pt}
\centering
\includegraphics[width=1\linewidth]{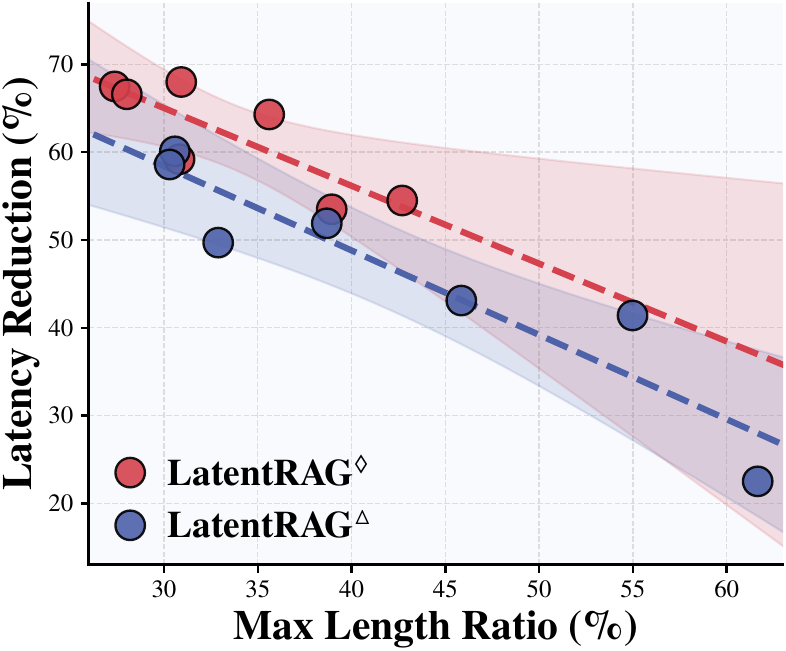}
\vspace{-15pt}
\caption{\textbf{Latency reduction using batch latent decoding vs. max length ratio.} Lower max length ratios are associated with higher latency reduction ratios. Each data point corresponds to the results on each dataset.}
\vspace{-15pt}
\label{fig:latency_vs_max_len_ratio}
\end{wrapfigure}

The efficiency gains from parallel decoding are more pronounced when sequence lengths are balanced, as this reduces padding overhead and avoids unnecessary computation. To characterize the impact of sequence length imbalance, we define the max length ratio as the ratio between the token count of the longest thought or subquery sequence and the total token count within a decoding batch. A higher max length ratio indicates a more imbalanced batch, where a single long sequence dominates most tokens. Such imbalance reduces the efficiency gains of parallel decoding due to increased padding overhead and more LLM forward passes. As shown in Fig.~\ref{fig:latency_vs_max_len_ratio}, the latency reduction percentage decreases as the max length ratio increases. This trend indicates that the effectiveness of parallel decoding is strongly associated with the degree of sequence length balance. Nevertheless, across different datasets with varying max length ratios, our method with latent decoding consistently achieves significant latency reductions, demonstrating the effectiveness of the parallel decoding strategy. 

\subsection{Detailed Stage-wise Latency Comparison}\label{app:latency_comparison}

Table~\ref{tab:stage_wise_latency} shows the detailed stage-wise latency breakdown when using the Qwen3-Embedding-0.6B retrieval model. Compared to naive single-step RAG, Search-R1 and AutoRefine introduce significant latency overhead. The average latency across all datasets is approximately 15$\times$ that of naive RAG. This overhead mainly comes from the thought and subquery generation stages, which together account for about 90\% of the total latency. In contrast, our method, trained on trajectories generated by Search-R1 and AutoRefine, significantly reduces the overall latency by approximately 90\% compared to the corresponding baseline.

\begin{table}[t]
\small{\caption{\textbf{Detailed stage-wise latency breakdowns (ms) using the Qwen3-Embedding-0.6B retrieval model.} \cell{red!20}{Red} and \cell{green!20}{green} highlights indicate the thought and subquery generation time of the baselines and LatentRAG, respectively. Search-R1 and AutoRefine incur significantly higher latency in the thought and subquery generation stages compared to our method.}\label{tab:stage_wise_latency}}
\centering
\setlength{\tabcolsep}{3pt}
\begin{adjustbox}{max width = 1.0\linewidth}
\begin{tabular}{lc|ccccccc|c}
\toprule
\textbf{Methods} & \textbf{Stages} & \textbf{NQ} & \textbf{TriviaQA} & \textbf{PopQA} & \textbf{HotpotQA} & \textbf{2wiki} & \textbf{Musique} & \textbf{Bamboogle} & \textbf{Average} \\
\midrule
\multirow{4}{*}{Naive RAG} & Prefill & 51 (11.21\%) & 50 (14.93\%) & 49 (16.63\%) & 51 (16.94\%) & 50 (12.68\%) & 49 (11.54\%) & 49 (15.47\%) & 50 (13.85\%) \\
& Retrieval & 93 (20.54\%) & 93 (27.88\%) & 90 (30.49\%) & 90 (29.79\%) & 91 (23.21\%) & 91 (21.50\%) & 91 (28.80\%) & 91 (25.41\%) \\
& Answer Gen. & 309 (68.24\%) & 191 (57.19\%) & 157 (52.88\%) & 160 (53.26\%) & 251 (64.11\%) & 283 (66.96\%) & 175 (55.73\%) & 218 (60.74\%) \\
\cmidrule{2-10}
& Total & 452 & 334 & 297 & 300 & 391 & 422 & 314 & 359 \\
\midrule
\midrule
\multirow{6}{*}{Search-R1$^{\lozenge}$} & Prefill & 101 (2.85\%) & 85 (1.64\%) & 87 (2.43\%) & 111 (1.68\%) & 131 (1.90\%) & 130 (1.89\%) & 108 (2.21\%) & 108 (2.00\%) \\
& Thought Gen. & \cellcolor{red!20} 2,453 (69.04\%) & \cellcolor{red!20} 3,919 (75.40\%) & \cellcolor{red!20} 2,676 (74.58\%) & \cellcolor{red!20} 5,110 (77.56\%) & \cellcolor{red!20} 5,309 (76.67\%) & \cellcolor{red!20} 5,105 (74.56\%) & \cellcolor{red!20} 3,571 (72.82\%) & \cellcolor{red!20} 4,021 (74.84\%) \\
& Subquery Gen. & \cellcolor{red!20} 650 (18.30\%) & \cellcolor{red!20} 701 (13.49\%) & \cellcolor{red!20} 488 (13.60\%) & \cellcolor{red!20} 942 (14.30\%) & \cellcolor{red!20} 946 (13.66\%) & \cellcolor{red!20} 1,112 (16.25\%) & \cellcolor{red!20} 779 (15.88\%) & \cellcolor{red!20} 803 (14.94\%) \\
& Retrieval & 166 (4.68\%) & 148 (2.84\%) & 154 (4.28\%) & 216 (3.28\%) & 282 (4.07\%) & 290 (4.23\%) & 212 (4.33\%) & 210 (3.90\%) \\
& Answer Gen. & 183 (5.14\%) & 345 (6.63\%) & 184 (5.12\%) & 210 (3.18\%) & 256 (3.70\%) & 210 (3.06\%) & 234 (4.77\%) & 231 (4.31\%) \\
\cmidrule{2-10}
& Total & 3,553 & 5,198 & 3,588 & 6,589 & 6,925 & 6,846 & 4,904 & 5,372 \\
\midrule
\multirow{6}{*}{\textbf{LatentRAG}$^{\lozenge}$} & Prefill & 95 (19.33\%) & 90 (18.85\%) & 86 (17.25\%) & 110 (17.52\%) & 123 (17.45\%) & 126 (17.26\%) & 104 (16.64\%) & 105 (17.67\%) \\
& Thought Gen. & \cellcolor{green!20} 84 (17.01\%) & \cellcolor{green!20} 81 (16.98\%) & \cellcolor{green!20} 84 (16.77\%) & \cellcolor{green!20} 103 (16.51\%) & \cellcolor{green!20} 113 (16.01\%) & \cellcolor{green!20} 116 (15.87\%) & \cellcolor{green!20} 102 (16.43\%) & \cellcolor{green!20} 98 (16.45\%) \\
& Subquery Gen. & \cellcolor{green!20} 102 (20.86\%) & \cellcolor{green!20} 100 (20.96\%) & \cellcolor{green!20} 103 (20.52\%) & \cellcolor{green!20} 144 (23.00\%) & \cellcolor{green!20} 169 (23.96\%) & \cellcolor{green!20} 176 (24.10\%) & \cellcolor{green!20} 138 (22.08\%) & \cellcolor{green!20} 133 (22.43\%) \\
& Retrieval & 96 (19.47\%) & 103 (21.52\%) & 110 (21.88\%) & 138 (22.06\%) & 159 (22.53\%) & 166 (22.66\%) & 129 (20.72\%) & 129 (21.65\%) \\
& Answer Gen. & 115 (23.32\%) & 104 (21.69\%) & 118 (23.59\%) & 131 (20.91\%) & 141 (20.05\%) & 147 (20.11\%) & 150 (24.12\%) & 129 (21.80\%) \\
\cmidrule{2-10}
& Total & 491\textbf{\textcolor{darkgreen}{(-86.2\%)}} & 478\textbf{\textcolor{darkgreen}{(-90.8\%)}} & 501\textbf{\textcolor{darkgreen}{(-86.0\%)}} & 626\textbf{\textcolor{darkgreen}{(-90.5\%)}} & 704\textbf{\textcolor{darkgreen}{(-89.8\%)}} & 730\textbf{\textcolor{darkgreen}{(-89.3\%)}} & 623\textbf{\textcolor{darkgreen}{(-87.3\%)}} & 593\textbf{\textcolor{darkgreen}{(-89.0\%)}} \\
\midrule
\midrule
\multirow{6}{*}{AutoRefine$^{\vartriangle}$} & Prefill & 87 (1.81\%) & 82 (1.94\%) & 83 (1.89\%) & 96 (1.80\%) & 115 (2.18\%) & 115 (2.07\%) & 99 (2.34\%) & 97 (2.00\%) \\
& Thought Gen. & \cellcolor{red!20} 3,858 (80.67\%) & \cellcolor{red!20} 3,063 (72.52\%) & \cellcolor{red!20} 3,478 (79.11\%) & \cellcolor{red!20} 3,987 (74.60\%) & \cellcolor{red!20} 3,678 (69.87\%) & \cellcolor{red!20} 3,879 (69.86\%) & \cellcolor{red!20} 2,955 (69.94\%) & \cellcolor{red!20} 3,557 (73.69\%) \\
& Subquery Gen. & \cellcolor{red!20} 525 (10.97\%) & \cellcolor{red!20} 777 (18.40\%) & \cellcolor{red!20} 512 (11.65\%) & \cellcolor{red!20} 890 (16.66\%) & \cellcolor{red!20} 1,013 (19.24\%) & \cellcolor{red!20} 1,119 (20.15\%) & \cellcolor{red!20} 757 (17.91\%) & \cellcolor{red!20} 799 (16.55\%) \\
& Retrieval & 127 (2.65\%) & 123 (2.92\%) & 141 (3.20\%) & 173 (3.23\%) & 223 (4.24\%) & 228 (4.10\%) & 190 (4.50\%) & 172 (3.56\%) \\
& Answer Gen. & 186 (3.90\%) & 178 (4.21\%) & 182 (4.14\%) & 198 (3.71\%) & 235 (4.46\%) & 213 (3.83\%) & 224 (5.30\%) & 202 (4.19\%) \\
\cmidrule{2-10}
& Total & 4,782 & 4,223 & 4,397 & 5,344 & 5,264 & 5,553 & 4,224 & 4,827 \\
\midrule
\multirow{6}{*}{\textbf{LatentRAG}$^{\vartriangle}$} & Prefill & 80 (19.45\%) & 77 (19.31\%) & 75 (17.65\%) & 93 (17.66\%) & 104 (17.07\%) & 107 (16.74\%) & 97 (16.74\%) & 90 (17.63\%) \\
& Thought Gen. & \cellcolor{green!20} 70 (17.05\%) & \cellcolor{green!20} 70 (17.46\%) & \cellcolor{green!20} 72 (17.09\%) & \cellcolor{green!20} 90 (16.98\%) & \cellcolor{green!20} 100 (16.53\%) & \cellcolor{green!20} 105 (16.42\%) & \cellcolor{green!20} 96 (16.53\%) & \cellcolor{green!20} 86 (16.80\%) \\
& Subquery Gen. & \cellcolor{green!20} 76 (18.67\%) & \cellcolor{green!20} 76 (18.98\%) & \cellcolor{green!20} 81 (19.10\%) & \cellcolor{green!20} 111 (21.09\%) & \cellcolor{green!20} 133 (21.90\%) & \cellcolor{green!20} 140 (21.86\%) & \cellcolor{green!20} 125 (21.41\%) & \cellcolor{green!20} 106 (20.67\%) \\
& Retrieval & 71 (17.37\%) & 72 (17.98\%) & 76 (17.94\%) & 104 (19.65\%) & 125 (20.56\%) & 145 (22.71\%) & 117 (20.07\%) & 101 (19.77\%) \\
& Answer Gen. & 112 (27.46\%) & 105 (26.26\%) & 119 (28.22\%) & 130 (24.63\%) & 145 (23.94\%) & 142 (22.27\%) & 147 (25.25\%) & 129 (25.12\%) \\
\cmidrule{2-10}
& Total & 409\textbf{\textcolor{darkgreen}{(-91.4\%)}} & 400\textbf{\textcolor{darkgreen}{(-90.5\%)}} & 422\textbf{\textcolor{darkgreen}{(-90.4\%)}} & 528\textbf{\textcolor{darkgreen}{(-90.1\%)}} & 607\textbf{\textcolor{darkgreen}{(-88.5\%)}} & 639\textbf{\textcolor{darkgreen}{(-88.5\%)}} & 581\textbf{\textcolor{darkgreen}{(-86.2\%)}} & 512\textbf{\textcolor{darkgreen}{(-89.4\%)}} \\
\bottomrule 
\end{tabular}
\end{adjustbox}
\end{table}

\subsection{Impact of Trajectory Quality on Model Performance}\label{app:trajectory_quality}

\begin{table}[t]
\small{\caption{\textbf{Effect of trajectory quality.} EM scores (\%)$^{\uparrow}$ are reported for LatentRAG$^{\lozenge}$-3B trained on trajectories generated by Search-R1$^{\lozenge}$ models of different sizes (3B, 7B, and 14B). \textbf{\textcolor{darkgreen}{Green}} values show gains obtained by training with trajectories from larger models, compared to the 3B setting.}\label{tab:effect_of_trajectory_quality}}
\centering
\setlength{\tabcolsep}{3pt}
\begin{adjustbox}{max width = 1.0\linewidth}
\begin{tabular}{l|ccccccc|c}
\toprule
\textbf{Methods} & \textbf{NQ} & \textbf{TriviaQA} & \textbf{PopQA} & \textbf{HotpotQA} & \textbf{2wiki} & \textbf{Musique} & \textbf{Bamboogle} & \textbf{Average} \\
\midrule
LatentRAG$^{\lozenge}$-3B/3B & 39.09 & 54.44 & 43.04 & 38.00 & 37.34 & 12.45 & 24.80 & 35.59 \\
LatentRAG$^{\lozenge}$-3B/7B & 43.99\textbf{\textcolor{darkgreen}{(+12.5\%)}} & 56.77\textbf{\textcolor{darkgreen}{(+4.3\%)}} & 45.62\textbf{\textcolor{darkgreen}{(+6.0\%)}} & 43.36\textbf{\textcolor{darkgreen}{(+14.1\%)}} & 41.80\textbf{\textcolor{darkgreen}{(+11.9\%)}} & 17.25\textbf{\textcolor{darkgreen}{(+38.6\%)}} & \textbf{37.60}\textbf{\textcolor{darkgreen}{(+51.6\%)}} & 40.91\textbf{\textcolor{darkgreen}{(+14.9\%)}} \\
LatentRAG$^{\lozenge}$-3B/14B & \textbf{44.90}\textbf{\textcolor{darkgreen}{(+14.9\%)}} & \textbf{57.37}\textbf{\textcolor{darkgreen}{(+5.4\%)}} & \textbf{45.90}\textbf{\textcolor{darkgreen}{(+6.6\%)}} & \textbf{44.23}\textbf{\textcolor{darkgreen}{(+16.4\%)}} & \textbf{44.86}\textbf{\textcolor{darkgreen}{(+20.1\%)}} & \textbf{17.54}\textbf{\textcolor{darkgreen}{(+40.9\%)}} & 32.80\textbf{\textcolor{darkgreen}{(+32.3\%)}} & \textbf{41.09}\textbf{\textcolor{darkgreen}{(+15.4\%)}} \\
\bottomrule 
\end{tabular}
\end{adjustbox}
\end{table}

To investigate the effect of trajectory quality on model performance, we train the same model using trajectories generated by LLMs of different sizes. As shown in Fig.~\ref{fig:model_size_scaling} in the main paper, larger LLMs consistently achieve better performance, suggesting that they tend to produce higher-quality interaction trajectories. Therefore, we use trajectories generated by Search-R1 based on Qwen2.5 models of different scales to train our method with the Qwen2.5-3B model. As shown in Table~\ref{tab:effect_of_trajectory_quality}, LatentRAG models trained on trajectories generated by the 7B and 14B models yield an average improvement of approximately 15\% over the variant trained on trajectories generated by the 3B model. These improvements indicate that our method benefits significantly from higher-quality training trajectories, highlighting the importance of the quality of the model used for trajectory generation.

\subsection{Frozen-Retriever Experiments}

In our main experiments in the main paper, the retrieval model is not frozen to maintain the general flexibility of our framework for adapting to the retrieval alignment objective. In this section, we explore whether latent representations can be understood by a text-based retrieval model when only the projector is optimized, without finetuning the retrieval model. This setting isolates the contribution of retrieval model finetuning and provides a more controlled comparison with the baselines.

\begin{table}[t]
\small{\caption{\textbf{Exact match score (\%) with and without retrieval model finetuning.} The full-finetuning and frozen-retriever variants of LatentRAG achieve similar performance, with limited or even no performance degradation compared to Search-R1.}
\label{tab:frozen_retriever_results}} \centering
\setlength{\tabcolsep}{6pt}
\begin{adjustbox}{max width = 1.0\linewidth}
\begin{tabular}{l|ccccccc|c}
\toprule
\textbf{Methods} & \textbf{NQ} & \textbf{TriviaQA} & \textbf{PopQA} & \textbf{HotpotQA} & \textbf{2wiki} & \textbf{Musique} & \textbf{Bamboogle} & \textbf{Average} \\
\midrule
\rowcolor{gray!30} \multicolumn{9}{c}{\textbf{Qwen3-Embedding-0.6B}} \\
\midrule
\textbf{Search-R1$^{\lozenge}$} & 43.93 & \textbf{61.70} & \textbf{47.31} & 43.05 & 43.27 & 19.61 & 38.40 & 42.47 \\
\textbf{LatentRAG$^{\lozenge}$} (full-finetuning) & \textbf{46.29} & 59.85 & 46.63 & \textbf{46.73} & \textbf{44.89} & 19.82 & 40.00 & 43.46 \textbf{\textcolor{darkgreen}{(+2.3\%)}} \\
\textbf{LatentRAG$^{\lozenge}$} (frozen-retriever) & 46.18 & 59.37 & 46.65 & 46.25 & 44.82 & \textbf{21.22} & \textbf{40.80} & \textbf{43.61} \textbf{\textcolor{darkgreen}{(+2.7\%)}} \\
\midrule
\rowcolor{gray!30} \multicolumn{9}{c}{\textbf{e5-base-v2}} \\
\midrule
\textbf{Search-R1$^{\lozenge}$} & 48.03 & \textbf{64.59} & \textbf{46.02} & 44.71 & \textbf{42.13} & \textbf{20.19} & \textbf{43.20} & \textbf{44.12} \\
\textbf{LatentRAG$^{\lozenge}$} (full-finetuning) & \textbf{49.03} & 61.30 & 43.04 & \textbf{46.41} & 38.61 & 18.87 & 39.20 & 42.35 \textbf{\textcolor{red}{(-4.0\%)}} \\
\textbf{LatentRAG$^{\lozenge}$} (frozen-retriever) & 48.37 & 61.11 & 42.65 & 45.77 & 38.92 & 18.29 & 40.80 & 42.27 \textbf{\textcolor{red}{(-4.2\%)}} \\
\midrule
\rowcolor{gray!30} \multicolumn{9}{c}{\textbf{jina-embeddings-v5-text-nano}} \\
\midrule
\textbf{Search-R1$^{\lozenge}$} & 45.79 & \textbf{63.16} & 46.54 & 44.31 & 42.95 & 20.40 & \textbf{44.80} & 43.99 \\
\textbf{LatentRAG$^{\lozenge}$} (full-finetuning) & 47.37 & 61.25 & \textbf{46.72} & \textbf{47.66} & 45.06 & \textbf{22.26} & 43.20 & \textbf{44.79} \textbf{\textcolor{darkgreen}{(+1.8\%)}} \\
\textbf{LatentRAG$^{\lozenge}$} (frozen-retriever) & \textbf{48.12} & 61.09 & 46.26 & 47.60 & \textbf{45.42} & 22.01 & 39.20 & 44.24 \textbf{\textcolor{darkgreen}{(+0.6\%)}} \\
\midrule
\rowcolor{gray!30} \multicolumn{9}{c}{\textbf{harrier-oss-v1-270m}} \\
\midrule
\textbf{Search-R1$^{\lozenge}$} & 44.40 & \textbf{62.80} & \textbf{47.14} & 44.88 & 44.16 & 18.62 & \textbf{40.00} & \textbf{43.14} \\
\textbf{LatentRAG$^{\lozenge}$} (full-finetuning) & \textbf{46.15} & 60.40 & 45.41 & \textbf{47.27} & \textbf{44.99} & \textbf{19.90} & 34.40 & 42.65 \textbf{\textcolor{red}{(-1.1\%)}} \\
\textbf{LatentRAG$^{\lozenge}$} (frozen-retriever) & 46.12 & 60.64 & 45.81 & \textbf{47.27} & 44.62 & 19.16 & 35.20 & 42.69 \textbf{\textcolor{red}{(-1.0\%)}} \\
\midrule
\rowcolor{gray!30} \multicolumn{9}{c}{\textbf{F2LLM-v2-330M}} \\
\midrule
\textbf{Search-R1$^{\lozenge}$} & 43.35 & \textbf{61.04} & \textbf{44.92} & 41.99 & 41.53 & \textbf{18.54} & \textbf{36.00} & 41.05 \\
\textbf{LatentRAG$^{\lozenge}$} (full-finetuning) & 45.54 & 58.56 & 44.21 & \textbf{44.58} & 43.02 & 18.37 & \textbf{36.00} & \textbf{41.47} \textbf{\textcolor{darkgreen}{(+1.0\%)}} \\
\textbf{LatentRAG$^{\lozenge}$} (frozen-retriever) & \textbf{46.29} & 59.40 & 44.87 & 44.38 & \textbf{43.59} & 18.20 & 32.00 & 41.25 \textbf{\textcolor{darkgreen}{(+0.5\%)}} \\
\bottomrule
\end{tabular}
\end{adjustbox}
\end{table}

As shown in Table~\ref{tab:frozen_retriever_results}, the frozen-retriever variant of LatentRAG achieves similar performance to the fully finetuned variant across all retrieval models evaluated, and both variants exhibit relative performance differences of less than 5\% compared to the Search-R1 baseline. This indicates that retrieval model finetuning is not a determining factor in the comparable performance of LatentRAG.

\subsection{Evaluation on Long-Horizon QA Tasks}

To evaluate whether LatentRAG can be applied to complex long-horizon QA tasks that require extensive multi-step reasoning and retrieval, we conduct additional experiments on two long-horizon QA benchmarks, Infoseek-Eval \citep{xia2026open} and BrowseComp-Plus \citep{chen2025browsecomp}, both of which provide curated corpora for constructing local retrieval systems.

\paragraph{Experimental Setup.} We use AgentCPM-4B \citep{chen2026agentcpm} as the teacher model, which is developed for long-horizon tasks. We construct training trajectories using the Infoseek-Train dataset \citep{xia2026open} and train LatentRAG based on the Qwen3-4B-Thinking-2507 model \citep{yang2025qwen3technicalreport}. We further train a text-based SFT model using the same training trajectories as LatentRAG, denoted as Text-SFT. Text-SFT follows the same design as LatentRAG, except that it operates in the text space rather than the latent space, which serves as a controlled baseline for isolating the effect of latent reasoning and retrieval. 

\paragraph{Metrics.} Following the evaluation protocol of BrowseComp-Plus, we report accuracy measured by Qwen3-32B as the judge model. We also report the cover exact match score (CEM), a rule-based metric that measures whether the ground-truth answer is covered by the generated answer. Since models on long-horizon QA tasks often generate lengthy answers with detailed explanations, answer generation can account for a substantial portion of the total latency, making the efficiency gains from multi-step retrieval less apparent. Therefore, in addition to average end-to-end latency, we report the time to first token (TTFT) as an additional efficiency metric, which measures the time from receiving a request to generating the first answer token. By excluding the subsequent answer generation phase, TTFT directly captures the efficiency gains from reducing reasoning and retrieval overhead. TTFT also better reflects the perceived waiting time before a user receives the response.

\begin{table}[t]
\small{\caption{\textbf{Long-horizon experiments.} LatentRAG achieves the best performance while significantly reducing latency and TTFT compared to the baselines.}
\label{tab:long_horizon_experiments}} \centering
\setlength{\tabcolsep}{3pt}
\begin{adjustbox}{max width = 1.0\linewidth}
\begin{tabular}{l|ccccc|ccccc}
\toprule
\multirow{2}{*}{\textbf{Methods}} & \multicolumn{5}{c|}{\textbf{Infoseek-Eval}} & \multicolumn{5}{c}{\textbf{BrowseComp-Plus}} \\
& Acc.(\%)$^{\uparrow}$ & CEM(\%)$^{\uparrow}$ & Lat.(ms)$^{\downarrow}$ & TTFT(ms)$^{\downarrow}$ & \# Steps & Acc.(\%)$^{\uparrow}$ & CEM(\%)$^{\uparrow}$ & Lat.(ms)$^{\downarrow}$ & TTFT(ms)$^{\downarrow}$ & \# Steps \\ 
\midrule
AgentCPM & 46.33 & 38.67 & 165.88 & 131.48 & 21.61 & 14.70 & 14.10 & 468.46 & 438.63 & 29.10 \\
Text-SFT & 44.00 & 38.33 & 155.38 & 114.70 & 21.25 & 14.94 & 13.13 & 355.12 & 314.41 & 28.98 \\
LatentRAG & 53.00 & 45.00 & 44.84 & 4.61 & 19.39 & 16.14 & 16.27 & 46.31 & 6.80 & 28.78 \\
\bottomrule
\end{tabular}
\end{adjustbox}
\end{table}

\paragraph{Results.} As shown in Table~\ref{tab:long_horizon_experiments}, all methods take about 20 steps on Infoseek-Eval and 30 steps on BrowseComp-Plus, demonstrating that these tasks require substantially more iterative reasoning and retrieval than the traditional QA benchmarks. LatentRAG achieves the best performance on both datasets while reducing the overall latency of AgentCPM by approximately 73\% on Infoseek-Eval and 90\% on BrowseComp-Plus. As LatentRAG primarily accelerates the reasoning and retrieval stages before answer generation, it achieves substantial TTFT reductions compared to AgentCPM, with a reduction of approximately 96\% on InfoSeek-Eval and 98\% on BrowseComp-Plus.

\subsection{Latency Comparison using Optimized Serving Frameworks}

In our main experiments in the main paper, we measure the latency using the standard forward pass implemented in Hugging Face Transformers\footnote{\url{https://huggingface.co/docs/transformers}}. Beyond the standard Hugging Face Transformers implementation, recent serving frameworks such as vLLM\footnote{\url{https://github.com/vllm-project/vllm}}, TensorRT-LLM\footnote{\url{https://github.com/NVIDIA/TensorRT-LLM}}, and SGLang\footnote{\url{https://github.com/sgl-project/sglang}} are widely adopted to optimize the inference process. However, existing serving frameworks are not optimized for the latent retrieval workflow and cannot be directly applied to our method. Specifically, vLLM and TensorRT-LLM do not support accessing hidden states, and SGLang only supports full hidden state extraction, which introduces unnecessary overhead since latent retrieval only requires hidden states corresponding to the special subquery tokens (around 2\% of the full sequence). Therefore, we evaluate latency with the standard HuggingFace Transformers forward pass for the main experiments to ensure a fair comparison, which is consistent with the evaluation protocols commonly adopted by previous works on latent reasoning \citep{hao2025training, he2025semcot}.

Apart from the standard decoding implementation, most serving frameworks also implement speculative decoding \citep{leviathan2023fast}, which is widely adopted to mitigate the autoregressive bottleneck and reduce latency in traditional text-based decoding. Speculative decoding incorporates a small draft model to quickly generate multiple candidate tokens, which are then verified by the target model in a single forward pass. However, this approach requires the draft model to closely match the target model to achieve a high acceptance rate. In complex agentic RAG scenarios, a smaller draft model may struggle to generate outputs that are consistent with the target model. Moreover, speculative decoding introduces additional computational overhead from draft generation and verification. In contrast, LatentRAG requires only a single forward pass per retrieval step without this additional overhead.

To further assess whether the latency advantage of LatentRAG persists under optimized serving, we additionally compare LatentRAG, without any serving-specific optimization, against the baseline method Search-R1 served with vLLM. This provides a conservative comparison for LatentRAG, as Search-R1 benefits from optimized serving while LatentRAG does not. Furthermore, beyond the standard inference implementation in vLLM, we also evaluate its speculative decoding implementation to investigate whether it can further reduce the latency gap between Search-R1 and LatentRAG. We evaluate speculative decoding with two draft models: the base Qwen2.5-0.5B model and a finetuned variant, denoted as SFT-Qwen2.5-0.5B, which is further finetuned on Search-R1 trajectories to improve the speculative acceptance rate. 

Our evaluation focuses on minimizing per-request latency under sufficient computational resources, rather than optimizing throughput under resource constraints through batching. While throughput under resource constraints can be improved by allocating additional computational resources, reducing per-request latency remains a more challenging problem due to the sequential dependencies inherent in autoregressive generation. Therefore, our evaluation focuses on the fundamental latency lower bound of individual requests under sufficient computational resources, with all experiments conducted at a batch size of 1. Nevertheless, LatentRAG also reduces inference computational costs and may offer throughput benefits. However, a rigorous evaluation of throughput requires integrating our method with advanced batch inference frameworks, which we leave as future work.

\begin{table}[t]
\small{\caption{\textbf{Latency comparison with vLLM.} P95 and P99 denote the 95th and 99th percentile tail latencies, respectively. LatentRAG, even without any serving optimization, achieves lower latency than Search-R1 with vLLM serving.}
\label{tab:latency_comparison_with_vllm}} \centering
\setlength{\tabcolsep}{3pt}
\begin{adjustbox}{max width = 1.0\linewidth}
\begin{tabular}{l|l|c|ccc}
\toprule
\textbf{Methods} & \textbf{Draft Model} & \textbf{\# Spec. Tokens} & \textbf{Latency (ms)}$^{\downarrow}$ & \textbf{P95 (ms)}$^{\downarrow}$ & \textbf{P99 (ms)}$^{\downarrow}$ \\
\midrule
LatentRAG (HF Transformers) & -- & -- & 593 (1.0$\times$) & 821 & 886 \\
Search-R1 (HF Transformers) & -- & -- & 5,372 (9.1$\times$) & 9,737 & 16,843 \\
\midrule
Search-R1 (vLLM) & -- & -- & 1,506 (2.5$\times$) & 2,454 & 2,899 \\
Search-R1 (vLLM + Spec.) & Qwen2.5-0.5B & 3 & 2,129 (3.6$\times$) & 3,222 & 3,733 \\
Search-R1 (vLLM + Spec.) & Qwen2.5-0.5B & 5 & 2,788 (4.7$\times$) & 4,279 & 4,742 \\
Search-R1 (vLLM + Spec.) & Qwen2.5-0.5B & 10 & 4,546 (7.7$\times$) & 6,737 & 7,408 \\
Search-R1 (vLLM + Spec.) & SFT-Qwen2.5-0.5B & 3 & 1,738 (2.9$\times$) & 2,850 & 3,337 \\
Search-R1 (vLLM + Spec.) & SFT-Qwen2.5-0.5B & 5 & 2,048 (3.5$\times$) & 3,425 & 3,912 \\
Search-R1 (vLLM + Spec.) & SFT-Qwen2.5-0.5B & 10 & 2,750 (4.6$\times$) & 4,597 & 5,490 \\
\bottomrule
\end{tabular}
\end{adjustbox}
\end{table}

As shown in Table~\ref{tab:latency_comparison_with_vllm}, Search-R1 with vLLM serving significantly reduces inference latency compared to the HF Transformers implementation of Search-R1. However, even without any serving optimization, LatentRAG still achieves lower latency than Search-R1 with vLLM serving. Furthermore, applying speculative decoding with Qwen2.5-0.5B as the draft model does not reduce Search-R1 latency in our setting. Instead, it increases latency due to the additional draft generation and verification. Using the finetuned model SFT-Qwen2.5-0.5B as the draft model improves the speculative acceptance rate (from 32.48\% to 69.61\% when using 10 speculative tokens), but the overall latency still underperforms the standard vLLM inference. The P95 and P99 tail latencies show similar trends, with LatentRAG achieving the lowest latency.

\subsection{Influence of Latent Token Numbers}\label{app:latent_token_numbers}

\begin{wrapfigure}{r}{0.5\textwidth}
\vspace{-13pt}
\centering
\includegraphics[width=1\linewidth]{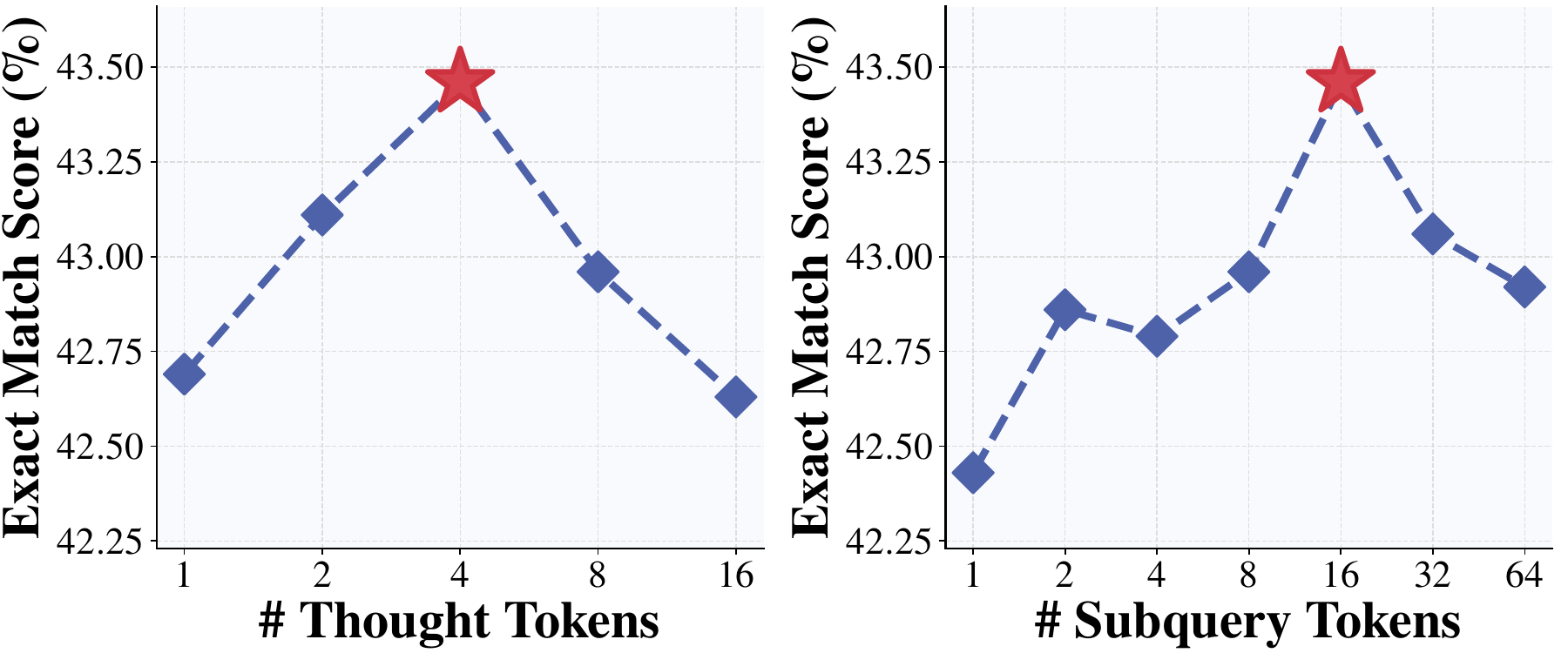}
\caption{\textbf{performance under different numbers of latent thought and subquery tokens.}}
\vspace{-10pt}
\label{fig:influence_of_latent_token_numbers}
\end{wrapfigure}

To investigate the impact of latent token numbers, we vary the number of latent thought tokens $m$ and the number of subquery tokens $n$ and evaluate the exact match scores under different configurations. As shown in Fig.~\ref{fig:influence_of_latent_token_numbers}, performance remains relatively stable across different settings. It increases slightly at first and reaches a peak when using 4 thought tokens and 16 subquery tokens for each step, followed by a decline as the token numbers continue to increase. This suggests that while additional latent tokens can provide more expressive capacity and increase performance, excessive tokens may introduce redundancy. Therefore, in our experiments, we set $m=4$ and $n=16$.

\subsection{Average Token Counts and Number of Forward Passes}\label{app:token counts}

\begin{table}[t]
\small{\caption{\textbf{Average token counts and number of forward passes per question.} \textbf{\textcolor{darkgreen}{(in)}} and \textbf{\textcolor{darkred}{(out)}} denote input and output tokens, respectively. Due to autoregressive token-by-token generation, output tokens incur more forward passes and thus higher latency.}\label{tab:token_counts}}
\centering
\setlength{\tabcolsep}{3pt}
\begin{adjustbox}{max width = 1.0\linewidth}
\begin{tabular}{l|c|c|c|c|c|c}
\toprule
\multirow{2}{*}{\textbf{Methods}} & \multicolumn{5}{c|}{\textbf{Token Counts}} & \multirow{2}{*}{\textbf{\# Forward Passes}} \\
& Thought & Subquery & Answer & Others & Total \\
\midrule
\textbf{Search-R1$^{\lozenge}$} & 121.8 \textbf{\textcolor{darkred}{(out)}} & 37.9 \textbf{\textcolor{darkred}{(out)}} & 9.6 \textbf{\textcolor{darkred}{(out)}} & 1325.5 \textbf{\textcolor{darkgreen}{(in)}} & 1325.5 \textbf{\textcolor{darkgreen}{(in)}} + 169.4 \textbf{\textcolor{darkred}{(out)}} & 169.4 \\
\textbf{LatentRAG$^{\lozenge}$} w/o decoding & 13.8 \textbf{\textcolor{darkgreen}{(in)}} & 39.0 \textbf{\textcolor{darkgreen}{(in)}} & 5.8 \textbf{\textcolor{darkred}{(out)}} & 1222.5 \textbf{\textcolor{darkgreen}{(in)}} & 1275.2 \textbf{\textcolor{darkgreen}{(in)}} + 5.8 \textbf{\textcolor{darkred}{(out)}} & \textbf{11.7} \\
\textbf{LatentRAG$^{\lozenge}$} w/ decoding & 13.8 \textbf{\textcolor{darkgreen}{(in)}} + 117.7 \textbf{\textcolor{darkred}{(out)}} & 39.0 \textbf{\textcolor{darkgreen}{(in)}} + 29.3 \textbf{\textcolor{darkred}{(out)}} & 5.8 \textbf{\textcolor{darkred}{(out)}} & 1436.2 \textbf{\textcolor{darkgreen}{(in)}} & 1489.0 \textbf{\textcolor{darkgreen}{(in)}} + 152.8 \textbf{\textcolor{darkred}{(out)}} & \textbf{52.8} \\
\bottomrule 
\end{tabular}
\end{adjustbox}
\end{table}

To analyze token usage efficiency, we report the average token counts per question. We distinguish between input and output tokens, as output tokens are generated autoregressively and cannot be fully parallelized, typically incurring higher latency and being more costly in practice. For example, in the OpenAI API pricing\footnote{\url{https://openai.com/api/pricing/}}, output tokens are typically priced about 6$\times$ higher than input tokens. We also report the number of forward passes per question. The number of forward passes corresponds to how many sequential LLM forward computations are required, which typically relates to the overall latency under sufficient hardware resources. Moreover, this latency cannot be easily reduced by simply scaling up GPU computational resources as it is fundamentally constrained by sequential dependencies in the generation process.

As shown in Table~\ref{tab:token_counts}, Search-R1 generates substantially more output tokens due to explicit thought and subquery generation, which in turn leads to a large number of LLM forward passes and explains its high latency reported in the main paper. In contrast, our method directly computes latent thought or subquery tokens by feeding a sequence of special tokens in parallel, which requires only a single forward pass per thought or subquery. As a result, without latent decoding, our method requires less than 5\% of the output tokens compared to Search-R1, significantly reducing the number of forward passes and thereby achieving substantially lower latency, as reported in the main paper. As an option to improve transparency at the cost of additional latency, latent decoding increases the number of output tokens in our method to a level comparable to Search-R1. However, since in our method the thought and subquery sequences across different steps are conditionally independent given the latent tokens, these sequences can be decoded in parallel, which significantly reduces the number of LLM forward passes. Moreover, the decoding process depends only on the latent tokens rather than attending to the full interaction history, which can further reduce computational overhead in practice. Consequently, even with a comparable number of output tokens, our method with latent decoding still only requires much fewer forward passes and achieves higher efficiency.

\subsection{Case Studies}\label{app:case_studies}

To qualitatively analyze the behavior of our method, we present several case studies of the reasoning and retrieval processes of LatentRAG.

\paragraph{Success case analysis.} As shown in Success Case~\ref{case:success_case_searchr1_compositional} and \ref{case:success_case_autorefine_compositional}, our method successfully learns the reasoning and retrieval patterns of the respective baseline models. Models trained on trajectories from different baselines generate thoughts and subqueries that are similar to those of the original models. For instance, the decoded thoughts of LatentRAG$^{\lozenge}$ are able to capture the refinement structure in the reasoning process of AutoRefine. In Success Case~\ref{case:success_case_searchr1_comparison}, although both models arrive at the correct answer after a sequence of reasoning and retrieval steps, they exhibit redundant retrieval in the final stage. This suggests that undesirable behaviors of the teacher model may also be learned by our method, highlighting the importance of trajectory quality as discussed in Appendix~\ref{app:trajectory_quality}. 

\paragraph{Failure case analysis.} As shown in Failure Case~\ref{case:failure_case_searchr1_1} and \ref{case:failure_case_searchr1_2}, although the reasoning and retrieval processes of our method are correct, the model sometimes fails to produce fully consistent outputs, leading to incorrect answers under exact match evaluation. This might indicate that latent representations facilitate the learning of abstract concepts but are less effective for precise lexical output. Nevertheless, our method maintains competitive performance while significantly reducing overall latency by 90\%, highlighting the value of latent reasoning and retrieval in agentic RAG. Future research can further investigate how to balance the use of latent representations and precise contextual information for accurate answer generation.

\begin{figure*}[h]
\begin{center}
\includegraphics[trim={0cm 0cm 0cm 0cm},clip,width=1.0\linewidth]{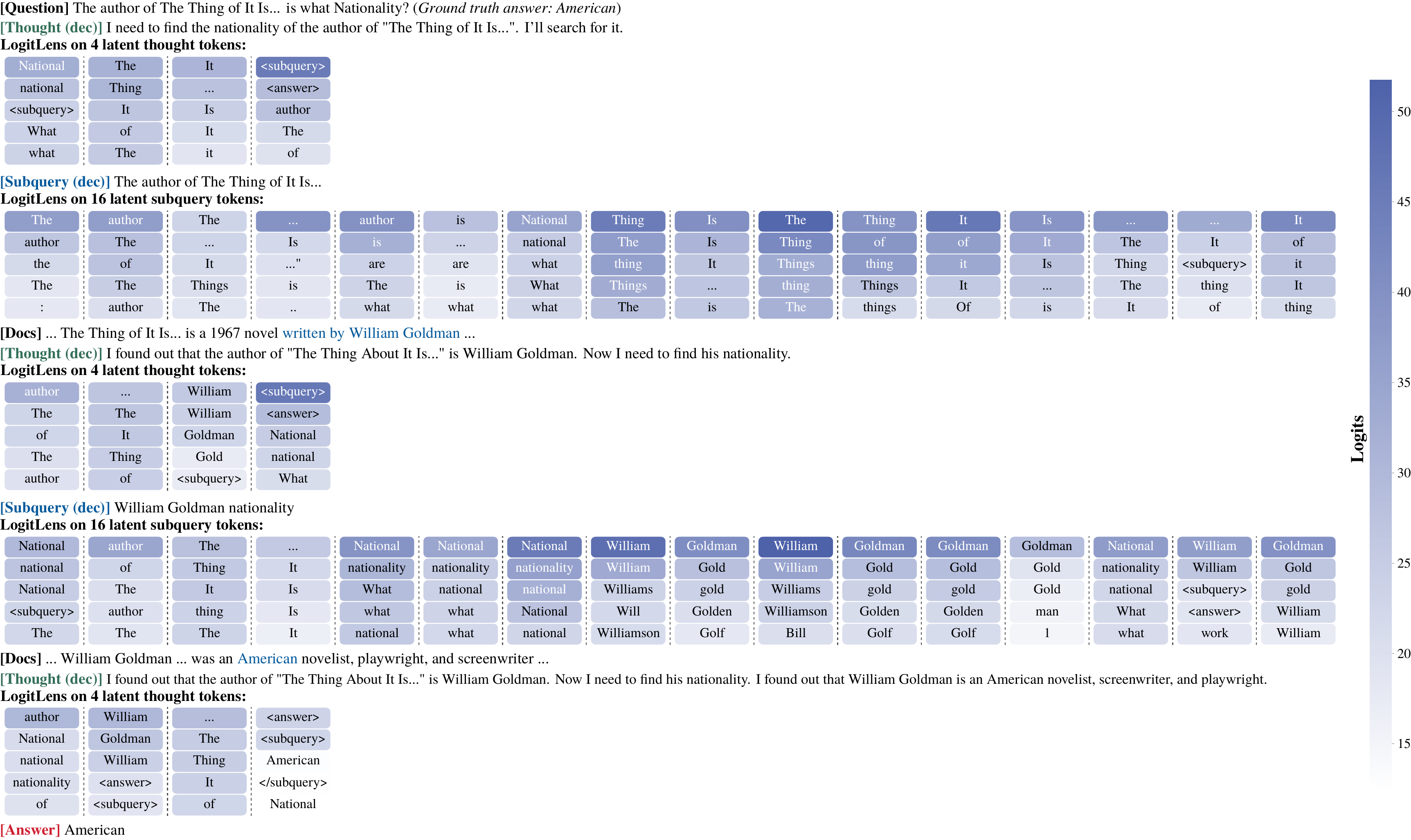}
\end{center}
\caption{\textbf{LogitLens Case Study 1 on LatentRAG$^{\lozenge}$.} Latent thought and subquery tokens in the first step align with tokens related to the first subquery, \textit{The author of The Thing of It Is...}, while those in the second step shift toward tokens related to the second subquery, \textit{William Goldman nationality}. A latent token can encode the whole semantic concept, such as \textit{The Thing of It Is...} or \textit{William Goldman}.}
\label{fig:latent_logits_case_1}
\end{figure*}

\begin{figure*}[h]
\begin{center}
\includegraphics[trim={0cm 0cm 0cm 0cm},clip,width=1.0\linewidth]{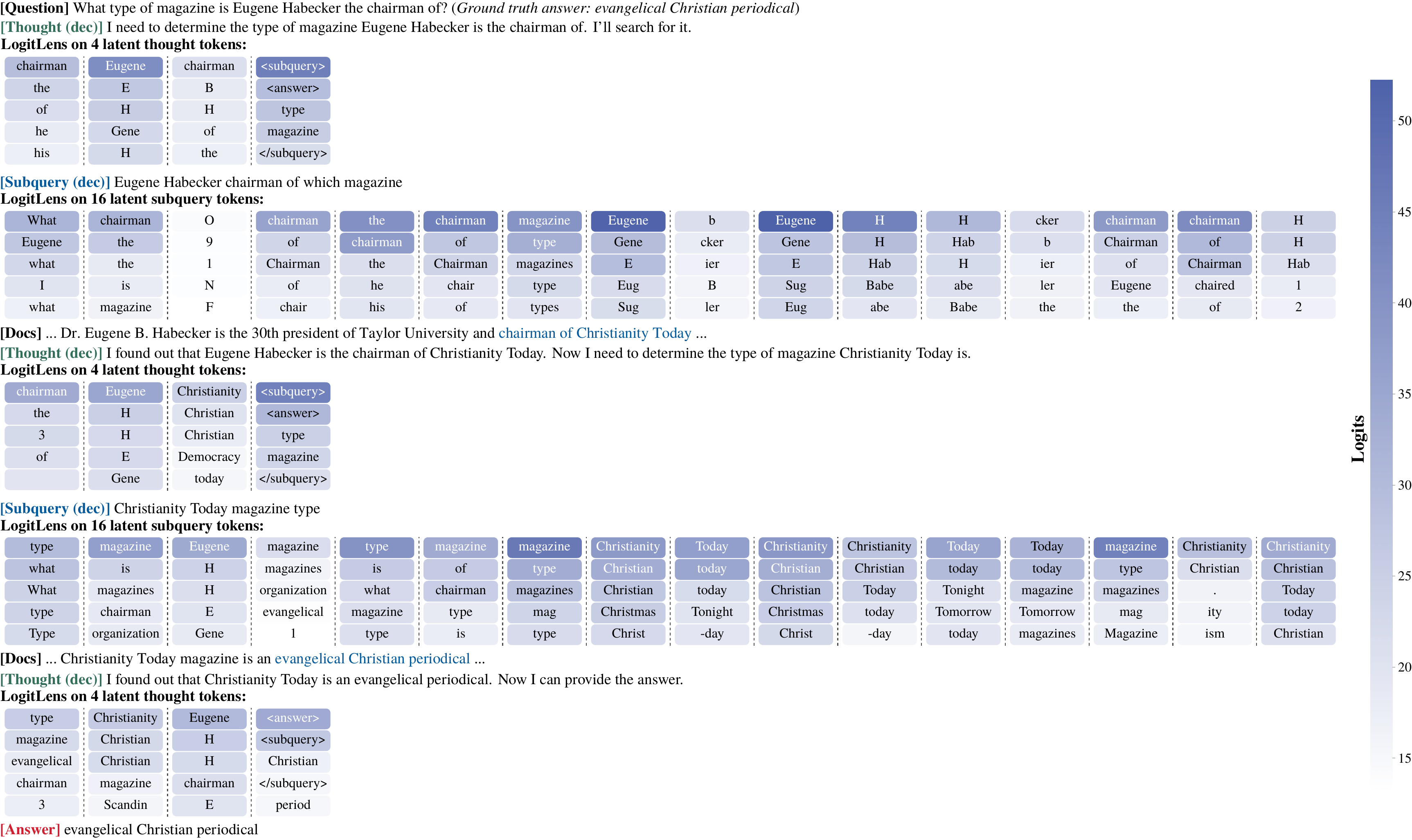}
\end{center}
\caption{\textbf{LogitLens Case Study 2 on LatentRAG$^{\lozenge}$.} Latent thought and subquery tokens in the first step align with tokens related to the first subquery, \textit{Eugene Habecker chairman of which magazine}, while those in the second step shift toward tokens related to the second subquery, \textit{Christianity Today magazine type}. A latent token can encode the whole semantic concept, such as \textit{magazine type} or \textit{Christianity Today}.}
\label{fig:latent_logits_case_2}
\vspace{-50pt}
\end{figure*}

\paragraph{LogitLens analysis.} To investigate what information is encoded in each latent thought or subquery token, we leverage LogitLens \citep{nostalgebraist2020logitlens} to analyze the generated latent tokens. LogitLens projects hidden states into the vocabulary space using the unembedding matrix of the LLM, enabling inspection of the token-level information encoded in the hidden states. Fig.~\ref{fig:latent_logits_case_1} \& \ref{fig:latent_logits_case_2} present the top-5 predicted language tokens by logits for each latent token. Although we do not explicitly constrain latent tokens to align with the LLM vocabulary space, the model still distributes these latent representations around semantically related vocabulary regions. In particular, the decoded vocabulary tokens from the thought and subquery tokens of the first step are closely related to the first subquery, while those from later steps gradually shift toward vocabulary regions associated with the second subquery and eventually the final answer. This progression indicates that the latent representations evolve as the reasoning process proceeds, with different latent tokens capturing information relevant to different stages of the reasoning and retrieval process. Additionally, unlike natural language tokenization, which typically represents text through a fixed subword decomposition that may split semantic units across multiple tokens, a latent token can encode the whole semantic concept, such as \textit{Christianity Today} or \textit{William Goldman}. These findings suggest that performing reasoning and retrieval in the latent space may provide more flexibility and expressivity than operating in natural language space.

{ 
\captionsetup[table]{name=Success Case, labelfont=bf, justification=raggedright, singlelinecheck=false}
\setcounter{table}{0}
\setlength{\textfloatsep}{6pt}
\setlength{\intextsep}{6pt}

\clearpage
\begin{table}[h]
\small{\caption{\textbf{Search-R1$^{\lozenge}$ vs. LatentRAG$^{\lozenge}$}}\label{case:success_case_searchr1_compositional}}
\centering
\setlength{\tabcolsep}{3pt}
\begin{adjustbox}{max width = 1.0\linewidth}
\begin{tabular}{lp{13.5cm}}
\toprule
\midrule
\textbf{[Question]} & What is the place of birth of Clara Novello's father? \textit{(Ground truth answer: London)} \\
\midrule
\midrule
\textbf{Search-R1$^{\lozenge}$: } & \\
\rowcolor{gray!10} \textbf{\textcolor{thought_color}{[Thought]}} & I need to find the place of birth of Clara Novello's father. I'll search for it. \\
\textbf{\textcolor{subquery_color}{[Subquery]}} & Clara Novello's father \\
\rowcolor{gray!10} \textbf{[Docs]} & \textbf{Chunk\_11245233 [Clara Novello]}: Ivor Novello carried on this family name. Clara Novello Clara Anastasia Novello (10 June 1818 – 12 March 1908) was an acclaimed soprano, \textcolor{subquery_color}{the fourth daughter of Vincent Novello}, a musician and music publisher, and his wife, Mary Sabilla Hehl. Her acclaimed soprano and pure style made her one of the greatest vocalists, alike in opera, oratorio and on the concert stage, from 1833 onwards. In 1843 she married Count Gigliucci, and retired in 1861. Charles Lamb wrote a poem ("To Clara N.") in her praise. She was born in Oxford Street, London, on 10 June 1818, the fourth daughter \newline ... \\
\textbf{\textcolor{thought_color}{[Thought]}} & I found out that Clara Novello's father is Vincent Novello. Now I need to find the place of birth of Vincent Novello. \\
\rowcolor{gray!10} \textbf{\textcolor{subquery_color}{[Subquery]}} & place of birth of Vincent Novello \\
\textbf{[Docs]} & \textbf{Chunk\_10809160 [Vincent Novello]}: Vincent Novello (6 September 1781 – 9 August 1861), English musician, son of an Italian who married an English wife, \textcolor{subquery_color}{was born in London}. He was a chorister and organist, but he is best known for bringing to England many works now considered major standards, and with his son he created a major publishing house. As a boy, Novello was a chorister at the Sardinian Embassy Chapel in Duke Street, Lincoln's Inn Fields, where he learnt the organ from Samuel Webbe; and from 1796 to 1822 he became in succession organist of the Sardinian, Spanish (in Manchester Square) and Portuguese \newline ... \\
\rowcolor{gray!10} \textbf{\textcolor{thought_color}{[Thought]}} & I found out that the place of birth of Vincent Novello is London. Now I can provide the answer. \\
\textbf{\textcolor{answer_color}{[Answer]}} & London \\

\midrule
\midrule

\textbf{LatentRAG$^{\lozenge}$:} & \\
\rowcolor{gray!10} \textbf{\textcolor{thought_color}{[Thought (dec)]}} & I need to find the place of birth of Clara Novello's father. I'll search for it. \\
\textbf{\textcolor{subquery_color}{[Subquery (dec)]}} & Clara Novello's father \\
\rowcolor{gray!10} \textbf{[Docs]} & \textbf{Chunk\_11245233 [Clara Novello]}: Ivor Novello carried on this family name. Clara Novello Clara Anastasia Novello (10 June 1818 – 12 March 1908) was an acclaimed soprano, \textcolor{subquery_color}{the fourth daughter of Vincent Novello}, a musician and music publisher, and his wife, Mary Sabilla Hehl. Her acclaimed soprano and pure style made her one of the greatest vocalists, alike in opera, oratorio and on the concert stage, from 1833 onwards. In 1843 she married Count Gigliucci, and retired in 1861. Charles Lamb wrote a poem ("To Clara N.") in her praise. She was born in Oxford Street, London, on 10 June 1818, the fourth daughter \newline ... \\
\textbf{\textcolor{thought_color}{[Thought (dec)]}} & I found out that Clara Novello's father is Vincent Novello. Now I need to find his place of birth. \\
\rowcolor{gray!10} \textbf{\textcolor{subquery_color}{[Subquery (dec)]}} & Vincent Novello place of birth \\
\textbf{[Docs]} & \textbf{Chunk\_10809160 [Vincent Novello]}: Vincent Novello (6 September 1781 – 9 August 1861), English musician, son of an Italian who married an English wife, \textcolor{subquery_color}{was born in London}. He was a chorister and organist, but he is best known for bringing to England many works now considered major standards, and with his son he created a major publishing house. As a boy, Novello was a chorister at the Sardinian Embassy Chapel in Duke Street, Lincoln's Inn Fields, where he learnt the organ from Samuel Webbe; and from 1796 to 1822 he became in succession organist of the Sardinian, Spanish (in Manchester Square) and Portuguese \newline ... \\
\rowcolor{gray!10} \textbf{\textcolor{thought_color}{[Thought (dec)]}} & I found out that the place of birth of Vincent Novello is London. Now I can provide the answer. \\
\textbf{\textcolor{answer_color}{[Answer]}} & London \\
\midrule
\bottomrule 
\end{tabular}
\end{adjustbox}
\end{table}

\clearpage
\begin{table}[h]
\small{\caption{\textbf{AutoRefine$^{\vartriangle}$ vs. LatentRAG$^{\vartriangle}$}}\label{case:success_case_autorefine_compositional}}
\centering
\setlength{\tabcolsep}{3pt}
\begin{adjustbox}{max width = 1.0\linewidth}
\begin{tabular}{lp{13.5cm}}
\toprule
\midrule
\textbf{[Question]} & What is the place of birth of Clara Novello's father? \textit{(Ground truth answer: London)} \\
\midrule
\midrule
\textbf{AutoRefine$^{\vartriangle}$: } & \\
\rowcolor{gray!10} \textbf{\textcolor{thought_color}{[Thought]}} & To answer this question, I need to find out who Clara Novello's father is and then determine his place of birth. \\
\textbf{\textcolor{subquery_color}{[Subquery]}} & Clara Novello's father \\
\rowcolor{gray!10} \textbf{[Docs]} & \textbf{Chunk\_11245233 [Clara Novello]}: Ivor Novello carried on this family name. Clara Novello Clara Anastasia Novello (10 June 1818 – 12 March 1908) was an acclaimed soprano, \textcolor{subquery_color}{the fourth daughter of Vincent Novello}, a musician and music publisher, and his wife, Mary Sabilla Hehl. Her acclaimed soprano and pure style made her one of the greatest vocalists, alike in opera, oratorio and on the concert stage, from 1833 onwards. In 1843 she married Count Gigliucci, and retired in 1861. Charles Lamb wrote a poem ("To Clara N.") in her praise. She was born in Oxford Street, London, on 10 June 1818, the fourth daughter \newline ... \\
\textbf{\textcolor{thought_color}{[Thought]}} & \textcolor{orange}{\texttt{<}refine\texttt{>}}From the documents, I learned that the father of Clara Novello was Vincent Novello. \textcolor{orange}{\texttt{<}/refine\texttt{>}} \\
\rowcolor{gray!10} \textbf{\textcolor{subquery_color}{[Subquery]}} & Vincent Novello's place of birth \\
\textbf{[Docs]} & \textbf{Chunk\_10809160 [Vincent Novello]}: Vincent Novello (6 September 1781 – 9 August 1861), English musician, son of an Italian who married an English wife, \textcolor{subquery_color}{was born in London}. He was a chorister and organist, but he is best known for bringing to England many works now considered major standards, and with his son he created a major publishing house. As a boy, Novello was a chorister at the Sardinian Embassy Chapel in Duke Street, Lincoln's Inn Fields, where he learnt the organ from Samuel Webbe; and from 1796 to 1822 he became in succession organist of the Sardinian, Spanish (in Manchester Square) and Portuguese \newline ... \\
\rowcolor{gray!10} \textbf{\textcolor{thought_color}{[Thought]}} & \textcolor{orange}{\texttt{<}refine\texttt{>}}From the documents, I learned that Vincent Novello was born in London. \textcolor{orange}{\texttt{<}/refine\texttt{>}} \\
\textbf{\textcolor{answer_color}{[Answer]}} & London \\

\midrule
\midrule

\textbf{LatentRAG$^{\vartriangle}$:} & \\
\rowcolor{gray!10} \textbf{\textcolor{thought_color}{[Thought (dec)]}} & To answer this question, I need to find out who Clara Novello's father is and then determine his place of birth. \\
\textbf{\textcolor{subquery_color}{[Subquery (dec)]}} & Clara Novello's father \\
\rowcolor{gray!10} \textbf{[Docs]} & \textbf{Chunk\_11245233 [Clara Novello]}: Ivor Novello carried on this family name. Clara Novello Clara Anastasia Novello (10 June 1818 – 12 March 1908) was an acclaimed soprano, \textcolor{subquery_color}{the fourth daughter of Vincent Novello}, a musician and music publisher, and his wife, Mary Sabilla Hehl. Her acclaimed soprano and pure style made her one of the greatest vocalists, alike in opera, oratorio and on the concert stage, from 1833 onwards. In 1843 she married Count Gigliucci, and retired in 1861. Charles Lamb wrote a poem ("To Clara N.") in her praise. She was born in Oxford Street, London, on 10 June 1818, the fourth daughter \newline ... \\
\textbf{\textcolor{thought_color}{[Thought (dec)]}} & \textcolor{orange}{\texttt{<}refine\texttt{>}}From the documents, I learned that the father of Clara Novello was Vincent Novello. \textcolor{orange}{\texttt{<}/refine\texttt{>}} \\
\rowcolor{gray!10} \textbf{\textcolor{subquery_color}{[Subquery (dec)]}} & What is the place of birth of Vincent Novello \\
\textbf{[Docs]} & \textbf{Chunk\_10809160 [Vincent Novello]}: Vincent Novello (6 September 1781 – 9 August 1861), English musician, son of an Italian who married an English wife, \textcolor{subquery_color}{was born in London}. He was a chorister and organist, but he is best known for bringing to England many works now considered major standards, and with his son he created a major publishing house. As a boy, Novello was a chorister at the Sardinian Embassy Chapel in Duke Street, Lincoln's Inn Fields, where he learnt the organ from Samuel Webbe; and from 1796 to 1822 he became in succession organist of the Sardinian, Spanish (in Manchester Square) and Portuguese \newline ... \\
\rowcolor{gray!10} \textbf{\textcolor{thought_color}{[Thought (dec)]}} & \textcolor{orange}{\texttt{<}refine\texttt{>}}From the documents, I learned that Vincent Novello was born in London. \textcolor{orange}{\texttt{<}/refine\texttt{>}} \\
\textbf{\textcolor{answer_color}{[Answer]}} & London \\
\midrule
\bottomrule 
\end{tabular}
\end{adjustbox}
\end{table}

\clearpage
\begin{table}[h]
\small{\caption{\textbf{Search-R1$^{\lozenge}$ vs. LatentRAG$^{\lozenge}$}}\label{case:success_case_searchr1_comparison}}
\centering
\setlength{\tabcolsep}{3pt}
\begin{adjustbox}{max width = 1.0\linewidth}
\begin{tabular}{lp{13.5cm}}
\toprule
\midrule
\textbf{[Question]} & Which film was released first, The Beginning Or The End or Hold Back The Night? \textit{(Ground truth answer: The Beginning or the End)} \\
\midrule
\midrule
\textbf{Search-R1$^{\lozenge}$: } & \\
\rowcolor{gray!10} \textbf{\textcolor{thought_color}{[Thought]}} & I need to determine which film, The Beginning or The End or Hold Back The Night, was released first. I'll search for the release years of both films. \\
\textbf{\textcolor{subquery_color}{[Subquery]}} & The Beginning or The End \\
\rowcolor{gray!10} \textbf{[Docs]} & \textbf{Chunk\_15135042 [The Beginning \textcolor{red}{Was} the End]}: The Beginning \textcolor{red}{Was} the End is a 1971 pseudo-scientific book written by ... \textcolor{red}{(Irrelevant retrieval results)} \newline ... \\
\textbf{\textcolor{thought_color}{[Thought]}} & I found out that \textcolor{red}{The Beginning or The End is a book}. Now I need to find the release year of the film Hold Back The Night and compare it to the release year of the film based on the book The Beginning or The End. \\
\rowcolor{gray!10} \textbf{\textcolor{subquery_color}{[Subquery]}} & Hold Back The Night release year \\
\textbf{[Docs]} & \textbf{Chunk\_15896613 [Hold Back the Night]}: \textcolor{subquery_color}{Hold Back the Night is a 1956 American war film} about the Korean War based on the 1951 novel by Pat Frank, who had been a war correspondent in Korea. The film was directed by Allan Dwan; his third film with John Payne and his third film about the United States Marine Corps, the others being "Abroad with Two Yanks" (1944) and "Sands of Iwo Jima" (1949). The film tells the story in flashbacks of a bottle of scotch carried by a World War II Marine lieutenant and Korean War captain, Sam MacKenzie. His girl Anne \newline ... \\
\rowcolor{gray!10} \textbf{\textcolor{thought_color}{[Thought]}} & I found out that Hold Back The Night is a film released in 1956. Now I need to find the release year of the film \textcolor{red}{based on the book} The Beginning or The End. \\
\textbf{\textcolor{subquery_color}{[Subquery]}} & The Beginning or The End release year \\
\rowcolor{gray!10} \textbf{[Docs]} & \textbf{Chunk\_10601403 [The Beginning or the End]}: seldom rises above cheery imbecility." According to MGM records, "The Beginning or the End" was made on a budget of \$2,632,000 (), but earned \$1,221,000 () in the United States and Canada and \$721,000 () elsewhere, resulting in a loss to the studio of \$1,596,000 (). \textcolor{subquery_color}{The Beginning or the End is a 1947 American docudrama film} about the development of the atomic bomb in World War II, directed by Norman Taurog, starring Brian Donlevy and Hume Cronyn, and released by Metro-Goldwyn-Mayer. The film dramatizes the creation of the atomic bomb in the Manhattan Project \newline .. \\
\textbf{\textcolor{thought_color}{[Thought]}} & I found out that The Beginning or the End is a film released in 1947. Now I can compare the release years of both films: \\
\rowcolor{gray!10} \textbf{\textcolor{subquery_color}{[Subquery]}} & Release year of Hold Back The Night \textcolor{red}{(Redundant retrieval)} \\
\textbf{[Docs]} & \textbf{Chunk\_15896613 [Hold Back the Night]}: ... \newline ... \\
\rowcolor{gray!10} \textbf{\textcolor{thought_color}{[Thought]}} & I found out that Hold Back The Night was released in 1956. Now I can compare it to the release year of The Beginning or the End, which I found to be 1947. \\
\textbf{\textcolor{answer_color}{[Answer]}} & The Beginning or the End \\

\midrule
\midrule

\textbf{LatentRAG$^{\lozenge}$:} & \\
\rowcolor{gray!10} \textbf{\textcolor{thought_color}{[Thought (dec)]}} & I need to determine which film, The Beginning or The End or Hold Back the Night, was released first. I'll search for the release years of both films. \\
\textbf{\textcolor{subquery_color}{[Subquery (dec)]}} & The Beginning Or The End \\
\rowcolor{gray!10} \textbf{[Docs]} & \textbf{Chunk\_10601403 [The Beginning or the End]}: seldom rises above cheery imbecility." According to MGM records, "The Beginning or the End" was made on a budget of \$2,632,000 (), but earned \$1,221,000 () in the United States and Canada and \$721,000 () elsewhere, resulting in a loss to the studio of \$1,596,000 (). \textcolor{subquery_color}{The Beginning or the End is a 1947 American docudrama film} about the development of the atomic bomb in World War II, directed by Norman Taurog, starring Brian Donlevy and Hume Cronyn, and released by Metro-Goldwyn-Mayer. The film dramatizes the creation of the atomic bomb in the Manhattan Project \newline .. \\
\textbf{\textcolor{thought_color}{[Thought (dec)]}} & I found out that The Beginning or The End is a 1947 American documentary film. Now I need to find the release year of Hold Back the Night. \\
\rowcolor{gray!10} \textbf{\textcolor{subquery_color}{[Subquery (dec)]}} & Hold Back The Night \\
\textbf{[Docs]} & \textbf{Chunk\_15896613 [Hold Back the Night]}: \textcolor{subquery_color}{Hold Back the Night is a 1956 American war film} about the Korean War based on the 1951 novel by Pat Frank, who had been a war correspondent in Korea. The film was directed by Allan Dwan; his third film with John Payne and his third film about the United States Marine Corps, the others being "Abroad with Two Yanks" (1944) and "Sands of Iwo Jima" (1949). The film tells the story in flashbacks of a bottle of scotch carried by a World War II Marine lieutenant and Korean War captain, Sam MacKenzie. His girl Anne \newline ... \\
\rowcolor{gray!10} \textbf{\textcolor{thought_color}{[Thought (dec)]}} & I found out that Hold Back the Night was released in 1956. Now I can compare the release years of both films. The Beginning or The End was released in 1947, and Hold Back the Night was released in 1956. \\
\textbf{\textcolor{subquery_color}{[Subquery (dec)]}} & release year of The Beginning \textcolor{red}{of} the End \textcolor{red}{(Redundant retrieval)} \\
\rowcolor{gray!10} \textbf{[Docs]} & \textbf{Chunk\_10601403 [The Beginning or the End]}: ... \newline .. \\
\textbf{\textcolor{thought_color}{[Thought (dec)]}} & I found out that The Beginning or The End (1947 film) was released in 1947. Now I can compare the release years of both films. \\
\rowcolor{gray!10} \textbf{\textcolor{answer_color}{[Answer]}} & The Beginning or the End \\
\midrule
\bottomrule 
\end{tabular}
\end{adjustbox}
\end{table}

} 

{ 
\captionsetup[table]{name=Failure Case, labelfont=bf, justification=raggedright, singlelinecheck=false}
\setcounter{table}{0}
\setlength{\textfloatsep}{0pt}
\setlength{\intextsep}{6pt}
\setlength{\floatsep}{0pt}

\clearpage

\begin{table}[h]
\small{\caption{\textbf{Search-R1$^{\lozenge}$ vs. LatentRAG$^{\lozenge}$}}\label{case:failure_case_searchr1_1}}
\centering
\setlength{\tabcolsep}{3pt}
\begin{adjustbox}{max width = 1.0\linewidth}
\begin{tabular}{lp{13.5cm}}
\toprule
\midrule
\textbf{[Question]} & Who lead the enquiry into the Iraq war? \textit{(Ground truth answer: Sir John Chilcot)} \\
\midrule
\midrule
\textbf{Search-R1$^{\lozenge}$: } & \\
\rowcolor{gray!10} \textbf{\textcolor{thought_color}{[Thought]}} & I need to find out who led the enquiry into the Iraq war. I'll search for it. \\
\textbf{\textcolor{subquery_color}{[Subquery]}} & who led the enquiry into the Iraq war \\
\rowcolor{gray!10} \textbf{[Docs]} & \textbf{Chunk\_6411881 [Iraq Inquiry]}: \textcolor{subquery_color}{The Iraq Inquiry (also referred to as the Chilcot Inquiry after its chairman, Sir John Chilcot)} was a British public inquiry into the nation's role in the Iraq War. The inquiry was announced in 2009 by Prime Minister Gordon Brown and published in 2016 with a public statement by Chilcot. On 6 July 2016, Sir John Chilcot announced the report's publication, more than seven years after the inquiry was announced. Usually referred to as the Chilcot report by the news media, the document stated that at the time of the invasion of Iraq in 2003, Saddam Hussein did \\
\textbf{\textcolor{thought_color}{[Thought]}} & I found out that the Iraq Inquiry, also referred to as the Chilcot Inquiry, was led by Sir John Chilcot. Now I can provide the answer. \newline ... \\
\rowcolor{gray!10} \textbf{\textcolor{answer_color}{[Answer]}} & Sir John Chilcot \\

\midrule
\midrule

\textbf{LatentRAG$^{\lozenge}$:} & \\
\rowcolor{gray!10} \textbf{\textcolor{thought_color}{[Thought (dec)]}} & I need to find out who led the inquiry into the Iraq War. I'll search for it. \\
\textbf{\textcolor{subquery_color}{[Subquery (dec)]}} & who led the inquiry into the Iraq War \\
\rowcolor{gray!10} \textbf{[Docs]} & \textbf{Chunk\_6411881 [Iraq Inquiry]}: \textcolor{subquery_color}{The Iraq Inquiry (also referred to as the Chilcot Inquiry after its chairman, Sir John Chilcot)} was a British public inquiry into the nation's role in the Iraq War. The inquiry was announced in 2009 by Prime Minister Gordon Brown and published in 2016 with a public statement by Chilcot. On 6 July 2016, Sir John Chilcot announced the report's publication, more than seven years after the inquiry was announced. Usually referred to as the Chilcot report by the news media, the document stated that at the time of the invasion of Iraq in 2003, Saddam Hussein did \newline ... \\
\textbf{\textcolor{thought_color}{[Thought (dec)]}} & I found out that the Iraq War was led by Sir John \textcolor{red}{Chilworth}. Now I can provide the answer. \\
\rowcolor{gray!10} \textbf{\textcolor{answer_color}{[Answer]}} & Sir John \textcolor{red}{Chilcott} \\
\midrule
\bottomrule 
\end{tabular}
\end{adjustbox}
\end{table}

\begin{table}[h]
\small{\caption{\textbf{Search-R1$^{\lozenge}$ vs. LatentRAG$^{\lozenge}$}}\label{case:failure_case_searchr1_2}}
\centering
\setlength{\tabcolsep}{3pt}
\begin{adjustbox}{max width = 1.0\linewidth}
\begin{tabular}{lp{13.5cm}}
\toprule
\midrule
\textbf{[Question]} & What was the name of the dog in the novel "Three Men in a Boat"? \textit{(Ground truth answer: Montmorency)} \\
\midrule
\midrule
\textbf{Search-R1$^{\lozenge}$: } & \\
\rowcolor{gray!10} \textbf{\textcolor{thought_color}{[Thought]}} & I need to find the name of the dog in the novel "Three Men in a Boat". I'll search for it. \\
\textbf{\textcolor{subquery_color}{[Subquery]}} & name of the dog in the novel "Three Men in a Boat" \\
\rowcolor{gray!10} \textbf{[Docs]} & \textbf{Chunk\_8650768 [Three Men in a Boat]}: is how undated it appears to modern readers – the jokes have been praised as fresh and witty. The three men are based on Jerome himself (the narrator Jerome K. Jerome) and two real-life friends, George Wingrave (who would become a senior manager at Barclays Bank) and Carl Hentschel (the founder of a London printing business, called Harris in the book), with whom Jerome often took boating trips. \textcolor{subquery_color}{The dog, Montmorency,} is entirely fictional but, "as Jerome admits, developed out of that area of inner consciousness which, in all Englishmen, contains an element of the dog". The trip is \newline ... \\
\textbf{\textcolor{thought_color}{[Thought]}} & I found out that the dog in the novel "Three Men in a Boat" is named Montmorency. Now I can provide the answer. \\
\rowcolor{gray!10} \textbf{\textcolor{answer_color}{[Answer]}} & Montmorency \\

\midrule
\midrule

\textbf{LatentRAG$^{\lozenge}$:} & \\
\rowcolor{gray!10} \textbf{\textcolor{thought_color}{[Thought (dec)]}} & I need to find the name of the dog in the novel "Three Men in a Boat". I'll search for it. \\
\textbf{\textcolor{subquery_color}{[Subquery (dec)]}} & what was the name of the dog in the novel "Three Men in a Boat" \\
\rowcolor{gray!10} \textbf{[Docs]} & \textbf{Chunk\_8650768 [Three Men in a Boat]}: is how undated it appears to modern readers – the jokes have been praised as fresh and witty. The three men are based on Jerome himself (the narrator Jerome K. Jerome) and two real-life friends, George Wingrave (who would become a senior manager at Barclays Bank) and Carl Hentschel (the founder of a London printing business, called Harris in the book), with whom Jerome often took boating trips. \textcolor{subquery_color}{The dog, Montmorency,} is entirely fictional but, "as Jerome admits, developed out of that area of inner consciousness which, in all Englishmen, contains an element of the dog". The trip is \newline ... \\
\textbf{\textcolor{thought_color}{[Thought (dec)]}} & I found out that the name of the dog in the novel "Three Men in a Boat" is \textcolor{red}{Montmorencie}. Now I can provide the answer. \\
\rowcolor{gray!10} \textbf{\textcolor{answer_color}{[Answer]}} & \textcolor{red}{Montmoreiras} \\
\midrule
\bottomrule 
\end{tabular}
\end{adjustbox}
\vspace{-20pt}
\end{table}

} 

\clearpage

\end{document}